\documentclass[acmsmall]{acmart}

\usepackage{lineno}  
\usepackage{booktabs}
\usepackage{url}

\usepackage{multirow}
\usepackage{color}

\newtheorem{upper bound}{Upper bound}

\usepackage{diagbox}
\usepackage{amsmath}
\usepackage{arydshln} 
\usepackage{color} 

\usepackage[linesnumbered,ruled]{algorithm2e} 

\SetAlFnt{\small}
\SetAlCapFnt{\small}
\SetAlCapNameFnt{\small}
\SetAlCapHSkip{0pt}
\IncMargin{-\parindent}

\usepackage{xspace}
\usepackage{multirow}
\usepackage{subfigure}

\setcopyright{rightsretained}

\acmJournal{CSUR}

\begin{document}
\title{A Survey of Implicit Discourse Relation Recognition}

\author{Wei Xiang}
\affiliation{%
	\institution{Huazhong University of Science and Technology}
	\streetaddress{Luoyu Road 1037}
	\city{WuHan}
	\state{Hubei}
	\country{China}
	\postcode{430074}
}
\email{xiangwei@hust.edu.com}

\author{Bang Wang}
\affiliation{%
	\institution{Huazhong University of Science and Technology}
	\streetaddress{Luoyu Road 1037}
	\city{WuHan}
	\state{Hubei}
	\country{China}
	\postcode{430074}
}
\email{wangbang@hust.edu.com}

\begin{abstract}
A discourse containing one or more sentences describes daily issues and events for people to communicate their thoughts and opinions. As sentences are normally consist of multiple text segments, correct understanding of the theme of a discourse should take into consideration of the relations in between text segments. Although sometimes a connective exists in raw texts for conveying relations, it is more often the cases that no connective exists in between two text segments but some implicit relation does exist in between them. The task of \textit{implicit discourse relation recognition} (IDRR) is to detect implicit relation and classify its sense between two text segments without a connective. Indeed, the IDRR task is important to diverse downstream \textit{natural language processing} tasks, such as text summarization, machine translation and so on. This article provides a comprehensive and up-to-date survey for the IDRR task. We first summarize the task definition and data sources widely used in the field. We categorize the main solution approaches for the IDRR task from the viewpoint of its development history. In each solution category, we present and analyze the most representative methods, including their origins, ideas, strengths and weaknesses. We also present performance comparisons for those solutions experimented on a public corpus with standard data processing procedures. Finally, we discuss future research directions for discourse relation analysis.
\end{abstract}




\renewcommand\shortauthors{Wei Xiang and Bang Wang}

\maketitle

\section{Introduction}\label{Sec:Introduction}
%研究的背景和意义
A discourse normally is referred to a sequence of clauses, sentences, or paragraphs in an article, which is often used to describe some issues, events, opinions, etc. Discourse parsing is to analyze the structure of the components of a discourse, which is a fundamental task in \textit{Natural Language Processing} (NLP)~\cite{Xue.N:et.al:2015:CoNLL, Xue.N:et.al:2016f:CoNLL}. It is widely agreed that a piece of text cannot be well understood in isolation, but should be related to its context~\cite{Lin.Z:et.al:2009:EMNLP}. \textit{Discourse relation recognition} (DRR), as a subtask of discourse parsing, aims at identifying the existence of some logical relation between two text segments in a discourse; And if existing, further classifies the relation sense into some predefined type, such as the \textsf{temporal}, \textsf{causal}, \textsf{contrastive} sense, etc~\cite{Pitler.E:et.al:2008:COLING}.

\par
Discourse relation recognition provides important information to many NLP tasks, such as question answering~\cite{Verberne.S:et.al:2007:SIGIR, Liakata.M:et.al:2013:EMNLP, Jansen.P:et.al:2014:ACL}, information extraction~\cite{Cimiano.P:et.al:2005:Elsevier, Li.H:et.al:2002:COLING}, machine translation~\cite{Li.J.J:et.al:2014:ACL, Guzman.F:et.al:2014:ACM, Meyer.T:Webber.B:2013:ACL-WS, Meyer.Popescu.B.A:2012:EACL-WS}, text summarization~\cite{Louis.A:et.al:2010:ACL, Yoshida.Y:et.al:2014:EMNLP, Gerani.S:et.al:2014:EMNLP}, sentiment analysis~\cite{Somasundaran.S:et.al:2009:EMNLP, Wang.C:Wang.B:2020:WWW, Wang.C:Wang.B:2020:NLPCC} and so on. For example, recognizing a causal relation between two text segments identifies the cause segment as well as the effect segment, which can assist in answering the '\textit{why}' question. A discourse may contain one or more \textit{connectives} in between text segments. A connective is a lexical unit (a word or a phrase) that may directly indicate some particular relation sense. For example, a connective\textit{'because'} often indicates a causal relation between two text segments. If a connective exists, it can be exploited for the DDR task. In such a case, the task is called \textit{explicit discourse relation recognition} (EDRR).

\par
On the other hand, it is not uncommon that two text segments express some relation sense but without any connective in between them. The task of detecting and classifying some latent relation in between two segments without an explicit connective is called \textit{implicit discourse relation recognition} (IDRR). Although it is very challenging, the IDRR task is of great importance in discourse parsing and other NLP tasks and has attracted lots of research efforts in recent years. This article surveys the developments, progresses and achievements in the IDRR community. Before that, we first introduce some basic definitions according to the widely acknowledged Penn Discourse TreeBank (PDTB) corpus~\cite{Miltsakaki.E:et.al:2004:LREC, Prasad.R:et.al:2008:LREC}.

%%%%%%%%%%%%%%%%%%%%%%%%%%%%%%%%%%%%%%%%%%%%%%%%%%%%%%%%%%%%%%%%%%%%%%%%%%%%%%%%%%%%%
%% Subsec: Discourse Relation Recognition Task
%%%%%%%%%%%%%%%%%%%%%%%%%%%%%%%%%%%%%%%%%%%%%%%%%%%%%%%%%%%%%%%%%%%%%%%%%%%%%%%%%%%%%
\subsection{Discourse Relation Recognition Task}
We use the following terminologies introduced in the PDTB corpus~\footnote{https://www.seas.upenn.edu/pdtb/}  (more details will be introduced in the next section)~\cite{Miltsakaki.E:et.al:2004:LREC, Prasad.R:et.al:2008:LREC}:

\begin{itemize}
	\item \textbf{Argument}: It is defined as a text segment in a discourse, containing at least one predicate for stating an issue, event, or opinion.
	\item \textbf{Sense}: It is the type of a discourse relation, such as the \textsf{comparison}, \textsf{contingency}, \textsf{expansion}, \textsf{temporal}, etc.
	\item \textbf{Connective}: It is a lexical item conveying some relation between two arguments, such as \textit{'but'}, \textit{'because of'}, etc.
\end{itemize}

\par
In the PDTB corpus,  the existence and sense of a relation between two arguments are judged and annotated by several professionals. The DRR task is to recognize such annotated relations by machine algorithms. Depending on whether a connective exists in the raw text, two kinds of DRR tasks are further defined. If a connective exists between two arguments, it corresponds to an EDRR task, whose main focus is to extract such an \textit{explicit connective} as a kind of relation trigger and classify it into some relation sense. If a connective does not exist in the raw text, it corresponds to an IDRR task, whose main focus is to detect some \textit{implicit} relation and classify its sense from the semantics and interactions of the two arguments. In the PDTB corpus, if two arguments holds an implicit relation, a connective, called as \textit{implicit connective}, is manually inserted into the raw text as an indication.

\par
\begin{figure}[h]
	\centering
	\includegraphics[width=0.7\linewidth]{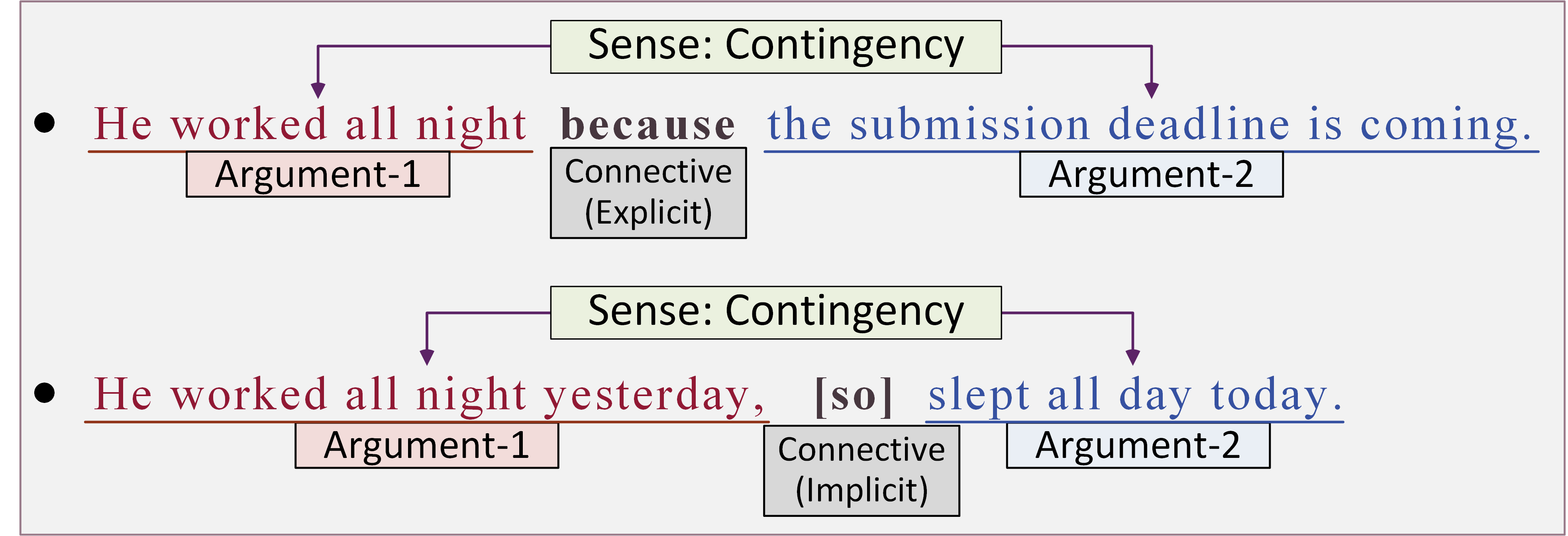}
	\caption{Examples of discourse relation annotation with explicit and implicit connectives in the PDTB corpus. The explicit connective is drawn from raw text, while the implicit connective is inserted by annotators.}
	\Description{Two sentences from PDTB corpus, each of them are annotated with two arguments, a predefined sense, and a connective. In the first sentence, the word 'because' is annotated as an explicit connective which exists in the raw text. In the second sentence, the word 'so' is annotated as an implicit connective which is inserted by annotators.}
	\label{Fig:DRRExample}
\end{figure}

Fig.~\ref{Fig:DRRExample} provides two annotation examples in the PDTB corpus, where text colors are used to denote arguments and connectives. In the first example, an explicit connective \textit{'because'} exists in the raw text, which is judged as revealing an explicit relation of \textsf{contingency} sense between the two arguments. In the second example, an implicit relation of  \textsf{contingency} is annotated for the two arguments, yet an implicit connective \textit{'so'} is also inserted by annotators.

%%%%%%%%%%%%%%%%%%%%%%%%%%%%%%%%%%%%%%%%%%%%%%%%%%%%%%%%%%%%%%%%%%%%%%%%%%%%%%%%%%%%%
%% Subsec: Summary of this Survey
%%%%%%%%%%%%%%%%%%%%%%%%%%%%%%%%%%%%%%%%%%%%%%%%%%%%%%%%%%%%%%%%%%%%%%%%%%%%%%%%%%%%%
\subsection{summary of this survey}
%本文简介
This article provides an up-to-date survey for the task of implicit discourse relation recognition. To the best of our knowledge, there has few similar review articles in this field~\cite{Atwell.K:et.al:2021:SIGDIAL, Hu.C:et.al:2020:ComputerSci., Yan.W:et.al:2016:Jour.OfChineseInfor.Proc.}. The Chinese survey by Yan et al. (2016)~\cite{Yan.W:et.al:2016:Jour.OfChineseInfor.Proc.} introduced the background, annotation, evaluation and challenge of the DRR task. Another Chinese survey by Hu et al. (2020)~\cite{Hu.C:et.al:2020:ComputerSci.} categorized some IDRR methods into argument encoding-based, argument interaction-based and semi-supervised. The most recent position paper by Atwell et al. (2021)~\cite{Atwell.K:et.al:2021:SIGDIAL} examined the performance of some discourse parsing models and discussed promising future directions.

\par
A few of articles on the IDRR performance comparison can be found~\cite{Braud.C:Denis.P:2015:EMNLP, Li.J.J:Nenkova.A:2016:ACL, Sun.K:Zhang.L:2018:JourOfQuantitativeLinguistics, Li.D:et.al:2019:Springer, Kim.N:et.al:2020:ACL, Liang.L:et.al:2020:ALC-WS}.
Braud and Denis (2015)~\cite{Braud.C:Denis.P:2015:EMNLP} presented a comparative framework for assessing the usefulness of various word representations in the IDRR task.
Li and Nenkova (2016)~\cite{Li.J.J:Nenkova.A:2016:ACL} analyzed the characterization factors for implicit instantiation relation detection in the PDTB corpus.
Sun and Zhang (2018)~\cite{Sun.K:Zhang.L:2018:JourOfQuantitativeLinguistics} compared the distribution pattern of discourse relations between the PDTB corpus and the \textit{Rhetorical Structure Theory Discourse Treebank} (RST-DT) corpus.
Li et al.(2019)~\cite{Li.D:et.al:2019:Springer} investigated the influence of neural components on the IDRR performance for several neural models.
Liang et al. (2020)~\cite{Liang.L:et.al:2020:ALC-WS} analyzed the additional annotations and relation sense distribution differences in the PDTB version 3.0, and Kim et al. (2020)~\cite{Kim.N:et.al:2020:ACL} presented an improved evaluation protocol considering the inconsistencies in published literatures.

%Kim et al. (2020)~\cite{Kim.N:et.al:2020:ACL} presented an improved evaluation protocol considering the inconsistencies in published literatures and reported baseline results on both PDTB 2.0 and 3.0.
%Liang et al. (2020)~\cite{Liang.L:et.al:2020:ALC-WS} analyzed the additional intra-sentential relation annotations and relation sense distributrion differences in PDTB 3.0.

\par
Compared with the aforementioned articles, we try to provide a comprehensive survey with systematic technique taxonomy for the IDRR task, not only providing its task definition, data sources and performance evaluations, but also categorizing the main approaches from the viewpoint of its development history. We present and analyze the most representative methods in each technique group, especially their origins, basics, advantages and disadvantages. We also discuss some promising directions for future research.

\par
In this article, we group the main approaches of the IDRR task into the traditional machine learning (ML) methods and deep learning (DL) methods. In the early years of discourse relation recognition, the common approach is to feed manually designed linguistic features into some traditional ML classifier to recognize relations. Recently, many deep learning methods have designed advanced neural networks to automatically conduct representation learning and classify argument-pair relations. In addition, we also introduce some semi-supervised learning methods that aim to solve the problem of limited resources of labelled data.

\par
The rest of this article is organized as follows: Section~\ref{Sec:Corpus} introduces the widely used PDTB corpus and the CoNLL dataset.We group the main approaches into the traditional machine learning methods in Section~\ref{Sec:Machine Learning} and deep learning methods in Section~\ref{Sec:DeepLearning Learning}. Section~\ref{Sec:Add Data} reviews the semi-supervised schemes in the literature. We also compare the performance for those algorithms experimented on the PDTB corpus in Section~\ref{Sec:Performance}. Finally, Section~\ref{Sec:Conclusion} concludes the survey with some discussions.

%%%%%%%%%%%%%%%%%%%%%%%%%%%%%%%%%%%%%%%%%%%%%%%%%%%%%%%%%%%%%%%%%%%%%%%%%%%%%%%%%%%%%
%%
%% Sec: Corpus and Shared Task
%%
%%%%%%%%%%%%%%%%%%%%%%%%%%%%%%%%%%%%%%%%%%%%%%%%%%%%%%%%%%%%%%%%%%%%%%%%%%%%%%%%%%%%%
\section{The PDTB corpus and the CoNLL dataset}\label{Sec:Corpus}
%本章内容简介
This section introduces two widely used data resources for discourse relation recognition: One is the PDTB corpus developed by the University of Pennsylvania~\cite{Miltsakaki.E:et.al:2004:LREC, Prasad.R:et.al:2008:LREC, Prasad.R:et.al:2007:UnivOfPenn, Webber.B:et.al:2019:UnivOfPenn}. Another is the CoNLL dataset released by the SIGNLL (ACL Special Interest Group on Natural Language Learning)~\footnote{https://www.signll.org/}. In both datasets, articles from news and journals are annotated with predefined discourse relation types, which can be regarded as with \textit{ground truth} labels.

%%%%%%%%%%%%%%%%%%%%%%%%%%%%%%%%%%%%%%%%%%%%%%%%%%%%%%%%%%%%%%%%%%%%%%%%%%%%%%%%%%%%%
%% Subsec: The PDTB Corpus
%%%%%%%%%%%%%%%%%%%%%%%%%%%%%%%%%%%%%%%%%%%%%%%%%%%%%%%%%%%%%%%%%%%%%%%%%%%%%%%%%%%%%
\subsection{The PDTB Corpus}
%PDTB语料的基本情况介绍，发布机构，文本来源，版本信息等。
The PDTB corpus is the largest one for the DRR task, which contains more than one million words of English texts from the Wall Street Journal on top of the Penn Treebank corpus~\cite{Marcus.M:et.al:1993:CL} and Propbank corpus~\cite{Palmer.M:et.al:2005:CL}, which have labeled arguments with a kind of predicate-argument structure~\cite{P.Marcus.M:et.al:1994:ARPA-HLT-Workshop, R.Kingsbury.P:Palmer.M:2002:LREC}. The first version PDTB 1.0 was released on April 2006 and a significantly extended version PDTB 2.0 was released on February 2008. The latest version PDTB 3.0 corpus has been released on March 2019 and updated on February 2020 through the Linguistic Data Consortium (LDC)~\footnote{http://www.ldc.upenn.edu}.
%\wbmark{on top of the Penn Treebank corpus~\cite{Marcus.M:et.al:1993:CL} and Propbank corpus~\cite{Palmer.M:et.al:2005:CL}, which has been widely used for part-of-speech (POS), syntactic structure and semantic analysis tasks~\cite{P.Marcus.M:et.al:1993:CL, P.Marcus.M:et.al:1994:ARPA-HLT-Workshop, R.Kingsbury.P:Palmer.M:2002:LREC}.}

\begin{figure}[b]
	\centering
	\includegraphics[width=0.9\linewidth]{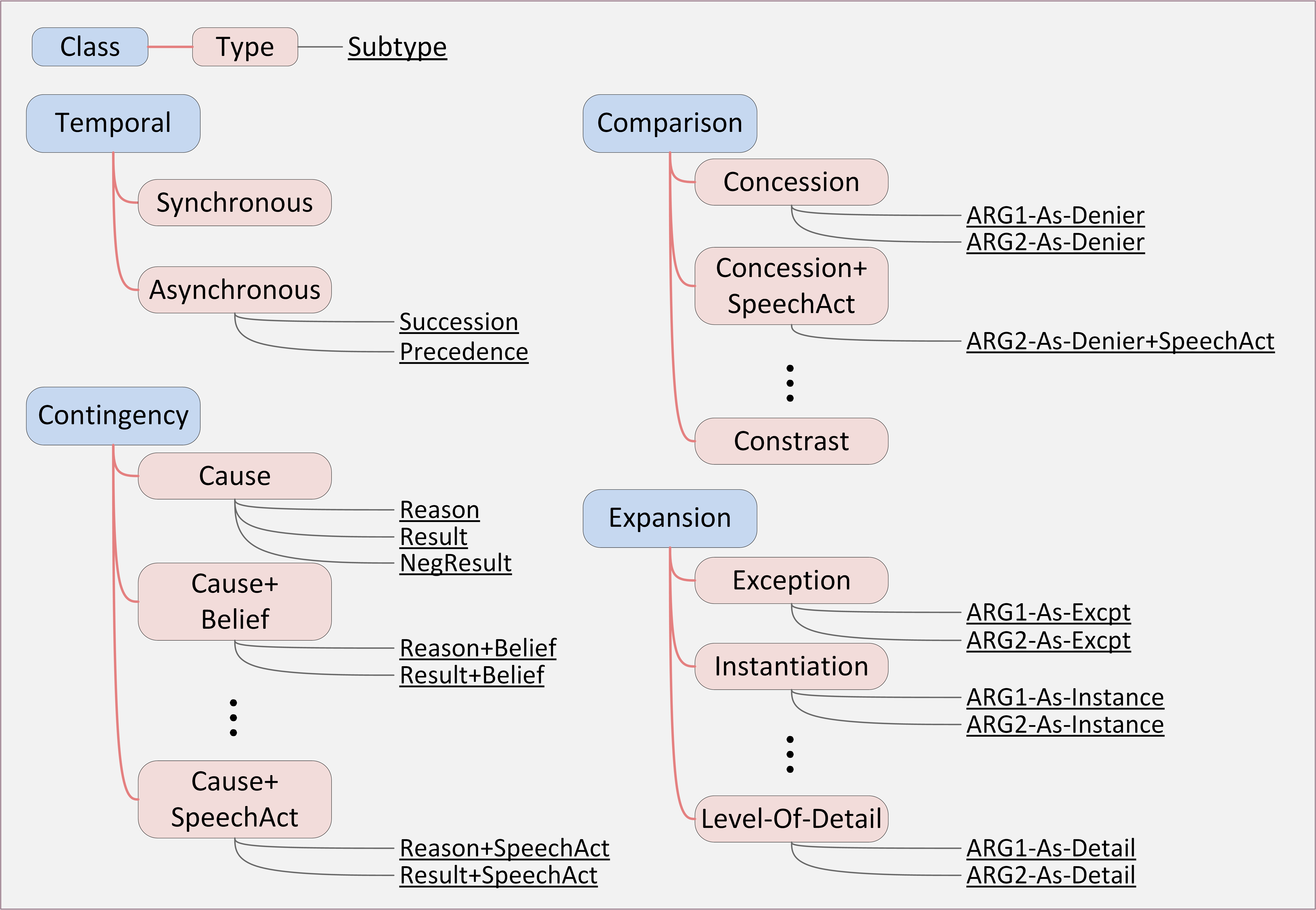}
	\caption{The hierarchy of sense annotation tags in the PDTB.}
	\Description{A sense annotation hierarchy tags, including three hierarchy: Class, Type, and Subtype. It has four classes at the top of the hierarchy: temporal, comparison, contingency and expansion, and several types and subtypes at the bottom.}
	\label{Fig:Sense}
\end{figure}

\par
Articles in the PDTB corpus are manually annotated by experts with domain knowledge. The PDTB corpus has annotated \textit{argument}, \textit{connective}, \textit{sense} for discourse relations, as well as \textit{attribution} and \textit{supplement} for arguments. Discourse relations in PDTB 1.0 and 2.0 are \textit{inter-sentential}, where arguments can only be a single clause, a single sentence, or a sequence of clauses and/or sentences. Some \textit{intra-sentential} discourse relations have been added to the latest PDTB version 3.0. Since arguments are annotated according to the minimality principle, any other span of text perceived to be relevant (but not necessary) to  the interpretation of arguments has been annotated as \textit{supplementary} information in parentheses.

%关系类型（type）
%\par
%The connective of discourse relation are broadly characterized into explicit and implicit, and the PDTB corpus also annotated three more kinds of connectives, i.e. Alternative Lexicalizations (AltLex), Entity Relation (EntRel) and Hypophora, that we take no account of.
%In PDTB, \textit{implicit discourse relation} is inferred by annotators, which implicitly expresses the relation between two adjacent argument in the absence of an explicit connective.
%In this case, inferred relation is annotated by manually inserting a connective called implicit connectives.
%For example, in the second discourse in Fig.~\ref{Fig:Example-1}, an implicit connective \textit{so} was inserted manually to express the inferred relation.

\par
\begin{figure}[b]
	\centering
	\includegraphics[width=0.9\linewidth]{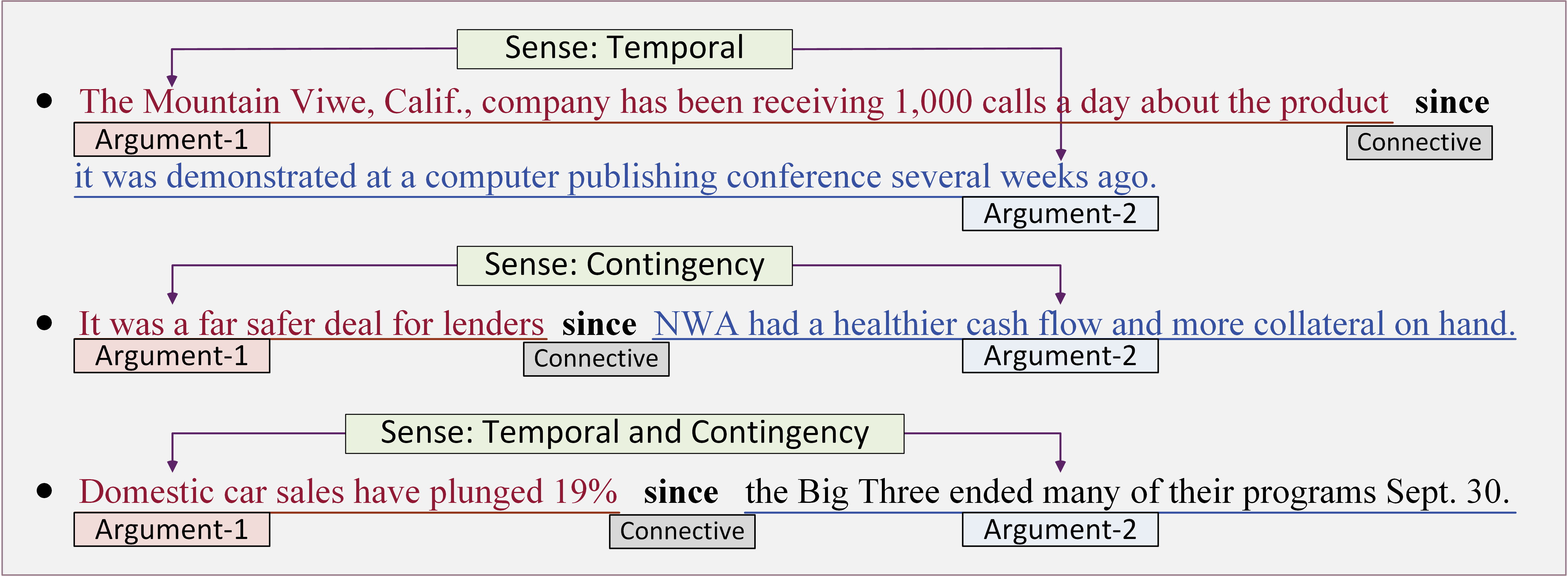}
	\caption{Examples of connective with multiple senses in the PDTB corpus.}
	\Description{Three sentences from PDTB corpus, each of them are annotated with two arguments, a same explicit connective 'since', and three different sense: Tmeporal, Contingency, and both Temporal and Contingency.}
	\label{Fig:Example-2}
\end{figure}

Each discourse relation between two consecutive arguments, called an \textit{argument-pair},  has been annotated with a sense; Where the sense annotation follows a hierarchical classification scheme with three levels: \textit{class}, \textit{type} and \textit{subtype}. Fig.~\ref{Fig:Sense} presents the hierarchy of sense annotation tags, including four classes: \textsf{temporal}, \textsf{comparison}, \textsf{contingency} and \textsf{expansion}, as well as several sense types and subtypes. Note that a discourse relation can be annotated with more than one sense in the PDTB corpus. Fig.~\ref{Fig:Example-2} presents such examples: In the first sentence, the connective \textit{'since'}  is annotated with a \textsf{temporal} relation sense; In the second sentence, the connective \textit{'since'} is annotated with a \textsf{contingency} relation sense. In the last one, it is with both the \textsf{contingency} and \textsf{temporal} sense. Multiple senses can also be inferred and annotated by inserting multiple implicit connectives.

% Attribution
%In addition, each discourse relation and its two arguments are annotated with key properties of \textbf{attribution} as their own features.

%\par
%% 统计数据
%In total, 40,600 tokens of discourse relation are annotated in PDTB-2.0, with 100 explicit connectives and 102 implicit connectives.
%

%%%%%%%%%%%%%%%%%%%%%%%%%%%%%%%%%%%%%%%%%%%%%%%%%%%%%%%%%%%%%%%%%%%%%%%%%%%%%%%%%%%%%
%% Subsec: The Shared Task
%%%%%%%%%%%%%%%%%%%%%%%%%%%%%%%%%%%%%%%%%%%%%%%%%%%%%%%%%%%%%%%%%%%%%%%%%%%%%%%%%%%%%
\subsection{The CoNLL Dataset and Shared Task}
%共享任务基本情况及所包含的子任务
The Conference on Natural Language Learning (CoNLL) is a yearly conference, in which a so-called Shared Task for \textit{Shallow Discourse Parsing} has been announced in the CoNLL-2015~\cite{Xue.N:et.al:2015:CoNLL} and CoNLL-2016~\cite{Xue.N:et.al:2016f:CoNLL}. The discourse relation in CoNLL is restrained between adjacent arguments, so it is called the \textit{shallow} discourse parsing. The CoNLL Shared Task introduces a slightly modified version of the PDTB corpus, which consists of the full PDTB dataset with some reductions of sense subtypes. In addition to containing all of the PDTB dataset, the CoNLL Shared Task includes an external blind test set and a second test set in the CoNLL-2015 and CoNLL-2016.

%The task of CoNLL relation recognition uses a piece of raw text as input and returns a discourse relation with a sense type and a discourse connective (explicit or implicit) among two arguments.

\par
The full CoNLL Shared Task consists of not only discourse relation recognition, but also discourse argument segmentation. For the explicit DRR task, connective detection and classification are also required. The tasks in the CoNLL focus on a kind of fine-grained discourse relation classification, i.e. classifying the lowest level of discourse relations in the discourse hierarchy as shown in Fig.~\ref{Fig:Sense}. Many models have been proposed for the CoNLL Shared Task, including some machine learning-based methods~\cite{Devi.S.L:et.al:2015:ACL-CoNLL, Chiarcos.C:Schenk.N:2015:ACL-CoNLL, Jian.P:et.al:2016:ACL-CoNLL, Kido.Y:Aizawa.A:2016:ACL-CoNLL, Li.Z:et.al:2016:ACL-CoNLL} and deep learning-based methods~\cite{Mihaylov.T:Frank.A:2016:ACL-CoNLL, Weiss.G:Bajec.M:2016:ACL-CoNLL, Li.J:et.al:2014:EMNLP, Wang.J:Lan.M:2016:ACL-CoNLL, Schenk.N:et.al:2016:ACL-CoNLL}.
%but only released to participants of the CoNLL-2015 conference.
%Such pipeline task is also called an end-to-end \textit{discourse relation parsing}:
%
%\par
%\begin{itemize}
%	\item \textbf{Connective identification:} locate explict connectives in a discourse.
%	\item \textbf{Argument segmentation:} identify the text spans in a discourse that serve as argument.
%	\item \textbf{Relation recognition:} predict the sense between two adjacent arguments in a discourse.
%\end{itemize}

%\par

%The CoNLL shared task also concerns other non-explicit connectives including \textit{EntRel} and \textit{AltLex}, in addition to the explicit and implicit connectives.

%%%%%%%%%%%%%%%%%%%%%%%%%%%%%%%%%%%%%%%%%%%%%%%%%%%%%%%%%%%%%%%%%%%%%%%%%%%%%%%%%%%%%
%%
%% Sec: Machine Learning Method
%%
%%%%%%%%%%%%%%%%%%%%%%%%%%%%%%%%%%%%%%%%%%%%%%%%%%%%%%%%%%%%%%%%%%%%%%%%%%%%%%%%%%%%%
\section{Implicit Discourse Relation Recognition based on Machine Learning}\label{Sec:Machine Learning}
The IDRR task can be straightforwardly formalized as an argument-pair classification problem, with the input of two arguments (and if necessary, other raw text in a discourse) and with the output of the classified relation sense between the two arguments. This section reviews those using traditional \textit{machine learning} (ML) algorithms for the IDRR task, like the \textit{Naive Bayes}, \textit{Maximum Entropy}, \textit{Support Vector Machine} (SVM) algorithm, etc. Those using neural network techniques (or so-called deep learning) will be reviewed in the next section. Note that they both are a kind of supervised learning approaches and require training data with ground truth labels.

%机器学习方法基本流程
\par
The basic idea and procedure are similar in those machine learning approaches. First, various features are designed to capture lexical, syntactic regularity and/or contextual information for an argument-pair. Next, each feature is represented by a numerical vector, mostly an \textit{one-hot} vector. These features are combined as the input of a ML classifier, which is trained based on training samples. Some ML approaches also enable \textit{feature selection} to select a subset of available features for improving classification performance. Finally, the trained classifier is applied to recognize the relation for an argument-pair, which returns a relation sense label with a confidence value indicating its classification likelihood.

\par
%\xwmod{[Delete this sentence:] Fig.~\ref{Fig:ML} presents the general procedure of ML-based IDRR classification.}
The key to the success of such ML classifiers is how to construct and select representative features, which is also called \textit{feature engineering}. For the IDRR task, linguistically informed features, such as lexical, syntactic, contextual features, have been widely used. Most of such features can be obtained via some open-source NLP tools, lexicons and annotated corpus, such as the Stanford CoreNLP~\footnote{https://stanfordnlp.github.io/CoreNLP/}, Levin English Verb Class~\footnote{https://www-personal.umich.edu/jlawler/levin.html}, General Inquirer Lexicon MPQA corpus~\footnote{https://www.wjh.harvard.edu/inquirer/inqdict.txt}. For feature selection, Park and Cardie~\cite{Park.J:Cardie.C:2012:SIGDIAL} and Li et al.~\cite{Li.S:et.al:2016:Jour.OfChineseInfor.Proc.} proposed a simple greedy approach: Starting with a single best-performing feature, in each iteration including the feature with the largest performance improvement. The iteration terminates until no significant improvement can be made. We next introduce some typical features widely used in the literature.

%\begin{figure*}[htbp]
%	\centering
%	\includegraphics[scale=0.8]{ML.png}
%	\caption{Illustration of implicit discourse relation recognition based on machine learning. Various linguistically informed features are designed and obtained via some source NLP tools, and then represented as vectors. Finally, the vectors are input to a classifier for implicit discourse relation classification and output the sense.}
%	\label{Fig:ML}
%\end{figure*}

%%%%%%%%%%%%%%%%%%%%%%%%%%%%%%%%%%%%%%%%%%%%%%%%%%%%%%%%%%%%%%%%%%%%%%%%%%%%%%%%%%%%%
%% linguistically informed features
%%%%%%%%%%%%%%%%%%%%%%%%%%%%%%%%%%%%%%%%%%%%%%%%%%%%%%%%%%%%%%%%%%%%%%%%%%%%%%%%%%%%%
\subsection{Lexical features}
The words of an argument-pair and their semantic characteristics can be utilized to design lexical features. Some commonly used lexical features include (1) \textit{word-pair}: a pair of two words in an argument-pair with the first word from the first argument and the second word from the second argument; (2) \textit{semantic tag}: some lexical characteristics of a word according to existing lexicons or corpora; (3) \textit{numerical value}: the number, dollar amounts, or percentage in arguments.

\par
\begin{figure}[h]
	\centering
	\includegraphics[scale=0.5]{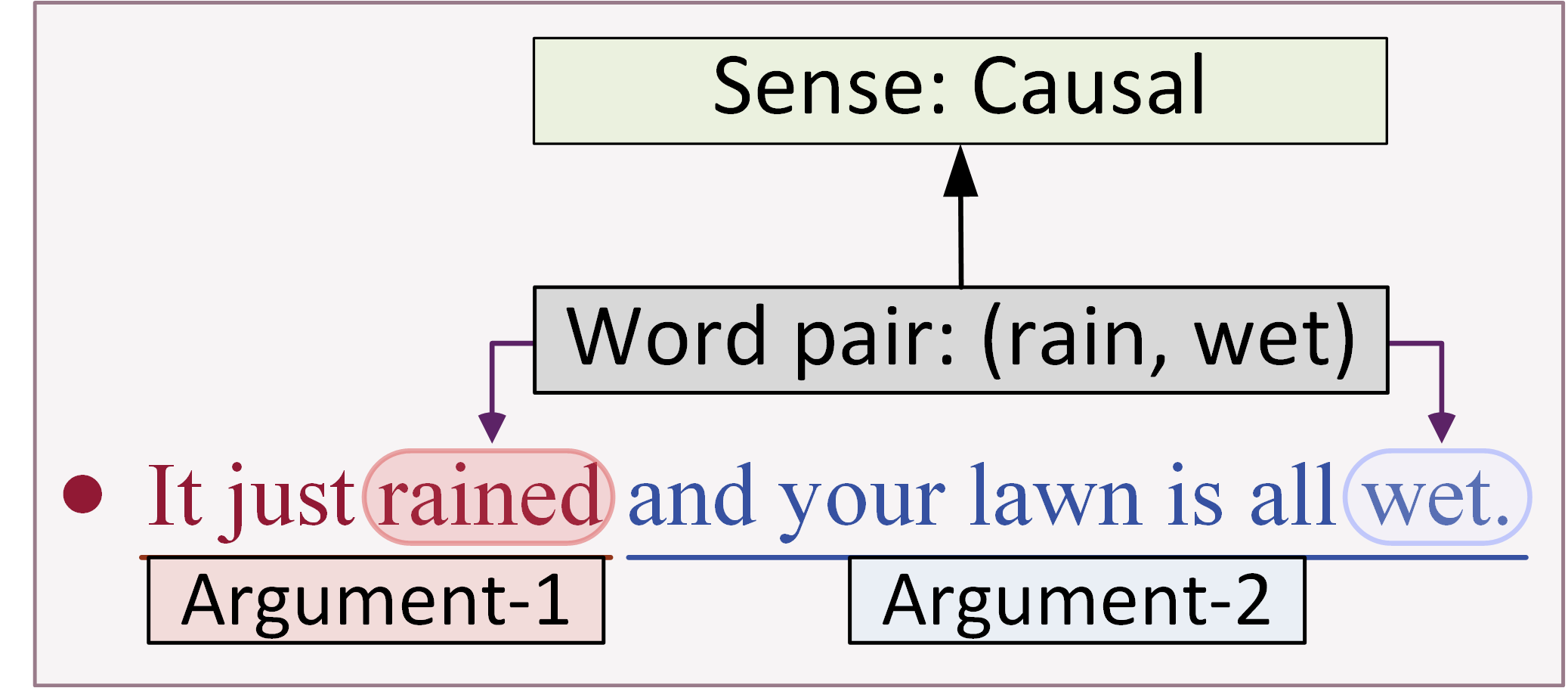}
	\caption{An example of word-pair which is indicative of a causal relation between two arguments.}
	\label{Fig:Word-pair}
	\Description{A sentence annotated with two arguments and a word pair 'rain, wet' indicating the sense of causal}
\end{figure}

The word-pair features could instinctively reveal some relation between arguments~\cite{Marcu.D:Echihabi.A:2002:ACL, Blair-Goldensohn.S:et.al:2007:ACL, Biran.O:McKeown.K:2013:ACL}. For example, a word-pair (\textit{rain, wet}) might be indicative of a causal relation between two arguments, as illustrated by Fig.~\ref{Fig:Word-pair}. Marcu and Echihabi~\cite{Marcu.D:Echihabi.A:2002:ACL} proposed to determine a discourse realtion $r_k$ between argument $A_1$ and $A_2$ by the word-pairs in terms of their cartesian product defined over the words in the two arguments ($w_i, w_j$) $\in$ $A_1 \times A_2$. They collected all word-pairs appearing around an explicit connective from a large corpus and computed the probability of each word-pair appearing in a specific relation sense. Biran and McKeown~\cite{Biran.O:McKeown.K:2013:ACL} found that using word-pairs may suffer from the data sparsity problem, as a word-pair in the training data may not appear in the test data. They aggregated some of the word-pairs with similar meaning to alleviate the data sparsity problem. Rutherford and Xue~\cite{Rutherford.A:Xue.N:2014:ACL} employed the Brown Cluster algorithm~\cite{Brown.P.F:et.al:1992:ACL} with a hierarchy word organization to provide a compact word representation for word-pairs.

%, where they represent the same relation sense as the corresponding connective. The probability of each word-pair appearing in a specific relation sense $P((w_i,w_j) \mid r_k)$ is computed by the maximum likelihood estimate, which are used as feature value.
%\xwmod{(???The most likely discourse relation holds between two arguments $A_1$ and $A_2$ is determined by taking the maximum over $argmax_{r_k} P(r_k \mid A_1, A_2)$.
	%According to Bayes Rule, it amounts to taking the maximum over $argmax_{r_k} [log P(A_1, A_2 \mid r_k)+logP(r_k)]$.
	%Assuming that each word-pair in the cartesian product is independent, $P(A_1, A_2 \mid r_k)$ is equivalent to $\prod_{(w_i, w_j) \in A_1 \times A_2}P((w_i,w_j) \mid r_k)$.)}

\par
A semantic tag marks a word with some predefined type according to existing lexicons or corpora, such as the sentiment polarity, verb class, inquirer, modality and so on~\cite{Pitler.E:et.al:2009:ACL}. For example, the Polarity Tag marks a word with a sentiment category, such as positive, negative, and neutral, according to the Multi-perspective Question Answering Opinion Corpus~\cite{Wilson.T:et.al:2005:ACL}. The Inquirer Tag marks each word with a more fine-grained semantic polarity according to the General Inquirer lexicon~\cite{Stone.P.J:et.al:1966:MIT}. The Verb Tag marks each verb in an argument with a particular class according to the Levin verb class~\cite{Levin.B:1993:ChicagoPress}. The Modality Tag indicates the presence of \textit{modal words} that are often used to express conditional statements, such as "\textit{can}", "\textit{may}", "\textit{should}", etc. A semantic tag feature is usually represented by its occurrences in arguments.
%The First-Last First3 feature uses the first, last, and first three words of each argument.

%词语特征其他方法
\par
Some other lexical features have also been designed for the IDRR task~\cite{Louis.A:et.al:2010:SIGDIAL, Roth.M:2017:IWCS, Lei.W:et.al:2018:AAAI, Li.S:et.al:2016:Jour.OfChineseInfor.Proc., Hong.Y:et.al:2019:FrontiersOfComp.Sci.}.
For example, Louis et al.~\cite{Louis.A:et.al:2010:SIGDIAL} found that about a quarter of all adjacent sentences are linked purely by the entity coherence, solely because they talk about the same entity. They designed entity category features, such as the entity grammatical role (an entity being the subject of a main clause or other clauses), the entity Part-of-Speech role (an entity being a pronoun, nominal, name or expletive), etc. Lei et al.~\cite{Lei.W:et.al:2018:AAAI} encoded topic continuity and attribution of an argument as features.
%Roth~\cite{Roth.M:2017:IWCS} introduced semantic roles feature based on FrameNet~\cite{Ruppenhoferand.J:et.al:2006:Inter.Comp.Sci.Ins.} and PropBank~\cite{Palmer.M:et.al:2005:CL}.

%Besides, Rutherford and Xue~\cite{Rutherford.A:Xue.N:2014:ACL} employed Brown cluster pair feature to represent discourse relation and incorporate coreference feature to identify the sense of implicit discourse relation.

\subsection{Syntactic features}
\begin{figure}[h]
	\centering
	\includegraphics[width=0.95\linewidth]{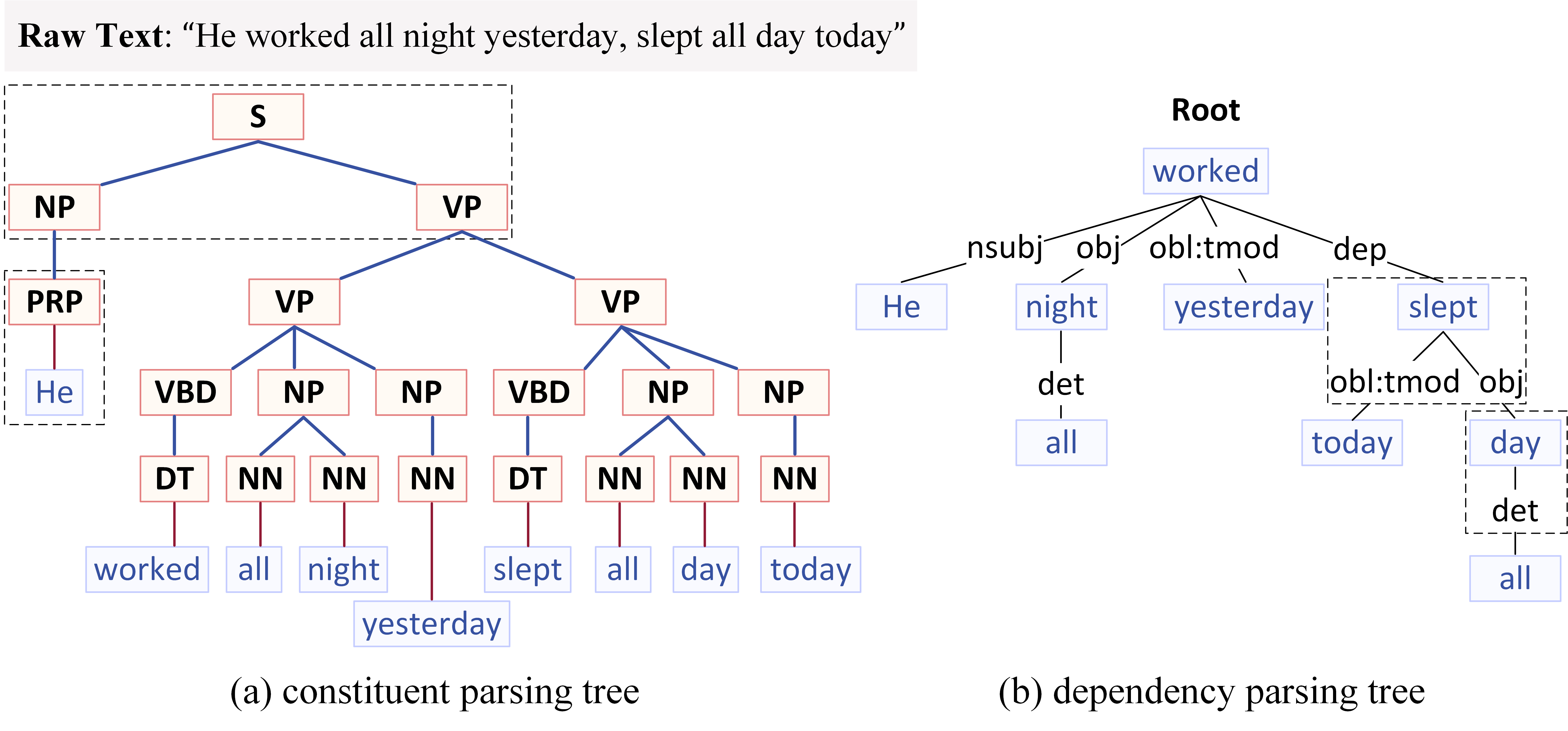}
	\caption{Examples of syntactic parsing tree obtained via Stanford Natural Language Parser\protect\footnotemark, in which figure (a) presents an example of constituent parsing tree and figure (b) presents an example of dependency parsing tree.}
	\Description{a constituent parsing tree and a dependency parsing tree of sentence: "He worked all night yesterday, slept all day today".}
	\label{Fig:Tree}
\end{figure}

Syntactic parsing is to analyze the grammatical structure of a sentence, which often uses a tree structure to describe the grammatical dependencies of the sentence components. Syntactic parsing can be divided into \textit{constituent parsing} and \textit{dependency parsing}. The constituent parsing aims at recognizing the phrase structures in a sentence and their hierarchy. Fig.~\ref{Fig:Tree} (a) presents an example of a constituent parsing tree: Each internal node represents a kind of phrase structure such as NP (Noun Phrase), VP (Vereb Phrase), etc.; Each leaf node represents a word in sentence with its Part-Of-Speech (POS) tag such as NN (Noun), VBD (Verb, past tense), PRP (Personal pronoun), DT (Determiner), etc.; The edges connect components of a phrase structure. The dependency parsing aims at recognizing the syntactic dependency between words in a sentence. Fig.~\ref{Fig:Tree} (b) presents an example of a dependency parsing tree, in which each node represents a word and the edges are syntactic dependency relations.

\footnotetext{https://nlp.stanford.edu/software/lex-parser.html}

\par
Syntactic knowledge has been shown effective and important for discourse analysis~\cite{LEE.A:et.al:2006:ProcOfTreebanksAndLinguisticTheories}.
Some researchers have proposed to integrate syntactic knowledge into feature engineering~\cite{Lin.Z:et.al:2009:EMNLP, Wang.W:et.al:2010:ACL, Li.J:Nenkova.A:2014:ACL, Li.H:et.al:2015:CCL}. For example, Lin et al.~\cite{Lin.Z:et.al:2009:EMNLP} extracted syntactic production rules from both the \textit{constituent parsing tree} (CPT) and \textit{dependency parsing tree} (DPT) of an argument. The syntactic production rule is the basic fragment of a syntactic tree. As illustrated by Fig.~\ref{Fig:Tree}, in CPT, it contains a head node and its dependent nodes, such as: ($S \to NP$, $S \to VP$), ($PRP \to$ "He"), etc.; While in DPT, it contains a head node and its dependency relation edges, such as: ("slept" $\to obl:tmod$, "slept" $\to  obj$), ("day" $\to det$), etc. All the syntactic production rules are collected from the training dataset. Then each syntactic production rule is represented as three binary features: whether this rule appears in the first argument, second argument and both arguments. In addition, Wang et al.~\cite{Wang.W:et.al:2010:ACL} proposed to use tree kernel-based approach to mine syntactic information from a constituent parsing tree; While the tree kernel measures the similarity between two structured objects.
%They further propose to leverage on temporal ordering information to constrain the interpretation of discourse relation.

\subsection{Contextual feautres}
Some implicit relation could appear immediately before or immediately after certain explicit relation~\cite{Pitler.E:et.al:2008:COLING}. Pitler et al.~\cite{Pitler.E:et.al:2009:ACL} defined contextual features as follows: For an implicit relation immediately preceding or following an explicit relation, include the explicit connective trigger and its sense as contextual features. Lin et al.~\cite{Lin.Z:et.al:2009:EMNLP} observed from the PDTB corpus that fully embedded arguments and shared arguments are the most common patterns. Specifically, between two adjacent discourse relations $r_1$ and $r_2$, if the previous discourse relation $r_1$ with its two arguments as a whole is one of the argument of the next discourse relation $r_2$, it is called a fully embedded argument. If an argument of previous discourse relation $r_1$ is also an argument of the next discourse relation $r_2$, it is called a shared argument. These fully embedded arguments and shared arguments between two adjacent discourse relations are represented as binary contextual features.

\par
\subsection{Implicit Connectives}\label{Sec:Implicit Connectives features}
%连接词在显式语篇关系识别中表现出好性能，可以用于隐式关系识别
%Some studies have shown that the presence of explicit connectives in a discourse can help improving implicit discourse relation recognition~\cite{Pitler.E:et.al:2008:COLING}.
%However, for implicit discourse relation, there are no connectives to explicitly mark the relation.
%Motivated by the critical importance of connectives in recognizing explicit discourse relations,
In the PDTB corpus, implicit connectives are manually inserted by annotators in between two arguments with implicit relations, although such implicit connectives are not present in the raw text. Some studies have proposed to exploit such manually inserted implicit connectives for the IDRR task~\cite{Zhou.Z:et.al:2010:COLING, Zhou.Z:et.al:2010:SIGDIAL, Zhou.X:et.al:2013:Jour.OfChineseInfor.Proc., Xu.Y:et.al:2012:IJCNN_IEEE, Hong.Y:et.al:2012:CIKM, Braud.C:Denis.P:2016:EMNLP, Braud.C:Denis.P:2016:EMNLP}. Zhou et al.~~\cite{Zhou.Z:et.al:2010:COLING} experimented that the F-score of implicit discourse relation classification can achieve 91.8\% on the PDTB corpus by simply mapping the annotated implicit connectives to their most frequent sense.
%Thus, it indicates that the prediction of implicit connectives can bring help to the discourse relation recognition.

%通过预测隐式关系连接词，然后根据预测的连接词判断关系
\par
In contrast to using hand-crafted features to directly recognize implicit relations, another interesting approach is to first predict an implicit connective and then recognize its relation sense as an explicit task. Zhou et al.~\cite{Zhou.Z:et.al:2010:COLING, Zhou.Z:et.al:2010:SIGDIAL} proposed to first predict implicit connectives with the use of a perplexity-based language model trained on a large amount of un-annotated corpora, and then to classify predicted implicit connectives to output relation senses.

\par
%直接用隐式连接词作为特征，输入模型，预测关系
Some studies have utilized annotated implicit connectives as additional features in ML models. Xu et al.~\cite{Xu.Y:et.al:2012:IJCNN_IEEE} predicted an implicit connective between two arguments using a ML model trained on various linguistically informed features. The predicted implicit connective is then used as an additional feature combined with other features as an input to a ML model again to output final implicit relation.
%Besides, Braud and Denis~\cite{Braud.C:Denis.P:2016:EMNLP} constructed a new distributional word representation based on large amounts of automatically extracted discourse connectives along with their arguments.

%通过在网络中搜索与当前“事件”对相似并包含连接词的文本，来判断当前的隐式关系
\par
Hong et al.~\cite{Hong.Y:et.al:2012:CIKM} and Zhou et al.~\cite{Zhou.X:et.al:2013:Jour.OfChineseInfor.Proc.} proposed a kind of cross-argument inference mechanism to infer implicit relations from a large number of comparable argument-pairs with explicit connectives. Such comparable argument-pairs are retrieved from the web via, e.g., Google search engine. This kind of pair-to-pair inference is based on the assumption that two argument-pairs with high content similarity would hold a same discourse relation.

%%%%%%%%%%%%%%%%%%%%%%%%%%%%%%%%%%%%%%%%%%%%%%%%%%%%%%%%%%%%%%%%%%%%%%%%%%%%%%%%%%%%%
%%
%% Sec: Deep Learning Method
%%
%%%%%%%%%%%%%%%%%%%%%%%%%%%%%%%%%%%%%%%%%%%%%%%%%%%%%%%%%%%%%%%%%%%%%%%%%%%%%%%%%%%%%
\section{Implicit Discourse Relation Recognition based on Deep Learning}\label{Sec:DeepLearning Learning}
%传统机器学习方法的不足
The aforementioned ML approaches heavily rely on various hand-crafted features to construct \textit{argument representation} for discourse relation recognition. Such features are normally designed and selected by professionals with linguistic knowledge and domain expertise; While the process of feature engineering is rather time-consuming and labor-intensive. Furthermore, conventional ML approaches might also suffer from the data sparsity problem due to their use of one-hot feature representations, which may not be able to well capture semantic and syntactic information in arguments.

\par
Recently, various neural networks have been proposed for automatic \textit{representation learning}, which use multiple layers of connected \textit{artificial neurons} to convert raw data embeddings into abstract yet informative representations. Compared with conventional machine learning, neural network-based deep learning does not require manual feature engineering and can learn to capture more useful information hidden in raw data. Indeed,  DL approaches have achieved significant improvements on many NLP tasks~\cite{Bengio.Y:et.al:2003:JourOfMachineLearningResearch, Manning.C:2015:Comput.Linguistics}, such as named entity recognition~\cite{Lample.G:et.al:2016:ACL, Zeng.D:et.al:2014:COLING, Nguyen.T:Grishman.R:2015:ACL-WsOnVectorSpaceModelingForNLP}, search query retrieval and question answering~\cite{Yih.W:et.al:2014:ACL, Shen.Y:et.al:2014:WWW}, sentence classification~\cite{Kim.Y:2014:EMNLP, Kalchbrenner.N:et.al:2014:ACL}, sentiment analysis~\cite{Wang.C:et.al:2019:Access, Wang.C:et.al:2019:SIGIR}, event extraction~\cite{Xiang.W:Wang.B:2019:Access}, and so on.

\par
For the IDRR task, recent years have also witnessed a boost of numerous neural models with significant performance improvements, compared with conventional ML approaches. Although those neural models employ diverse architectures with different types of neurons, connections and operations, they are with the same design objective, that is, how to learn to represent the most useful information in arguments, like semantic, syntactic, contextual information, as well as interactions between arguments. In the rest of this section, we review some representative types of neural networks for the IDRR task.

\par
Before reviewing neural models, we first introduce some common procedures adopted in these neural models. A piece of raw text is first divided into individual characters/words, each using a low-dimensional and real-valued vector as its representation, viz., \textit{word embedding}. Word embeddings can be pre-trained from external corpus via, say for example, the \textit{continuous bag-of-words model}~\cite{Mikolov.T:et.al:2013:arXiv} or the \textit{continuous skip-gram model}~\cite{Mikolov.T:et.al:2013:NeurIPS}. Some common techniques have been used for neural network training, including the activation function, back-propagation gradient, normalization, and dropout regularization, etc.

%%%%%%%%%%%%%%%%%%%%%%%%%%%%%%%%%%%%%%%%%%%%%%%%%%%%%%%%%%%%%%%%%%%%%%%%%%%%%%%%%%%%%
%% Subsec: Convolutional Neural Networks
%%%%%%%%%%%%%%%%%%%%%%%%%%%%%%%%%%%%%%%%%%%%%%%%%%%%%%%%%%%%%%%%%%%%%%%%%%%%%%%%%%%%%
\subsection{Convolutional Neural Networks}

\begin{figure}[h]
	\centering
	\includegraphics[width=0.8\linewidth]{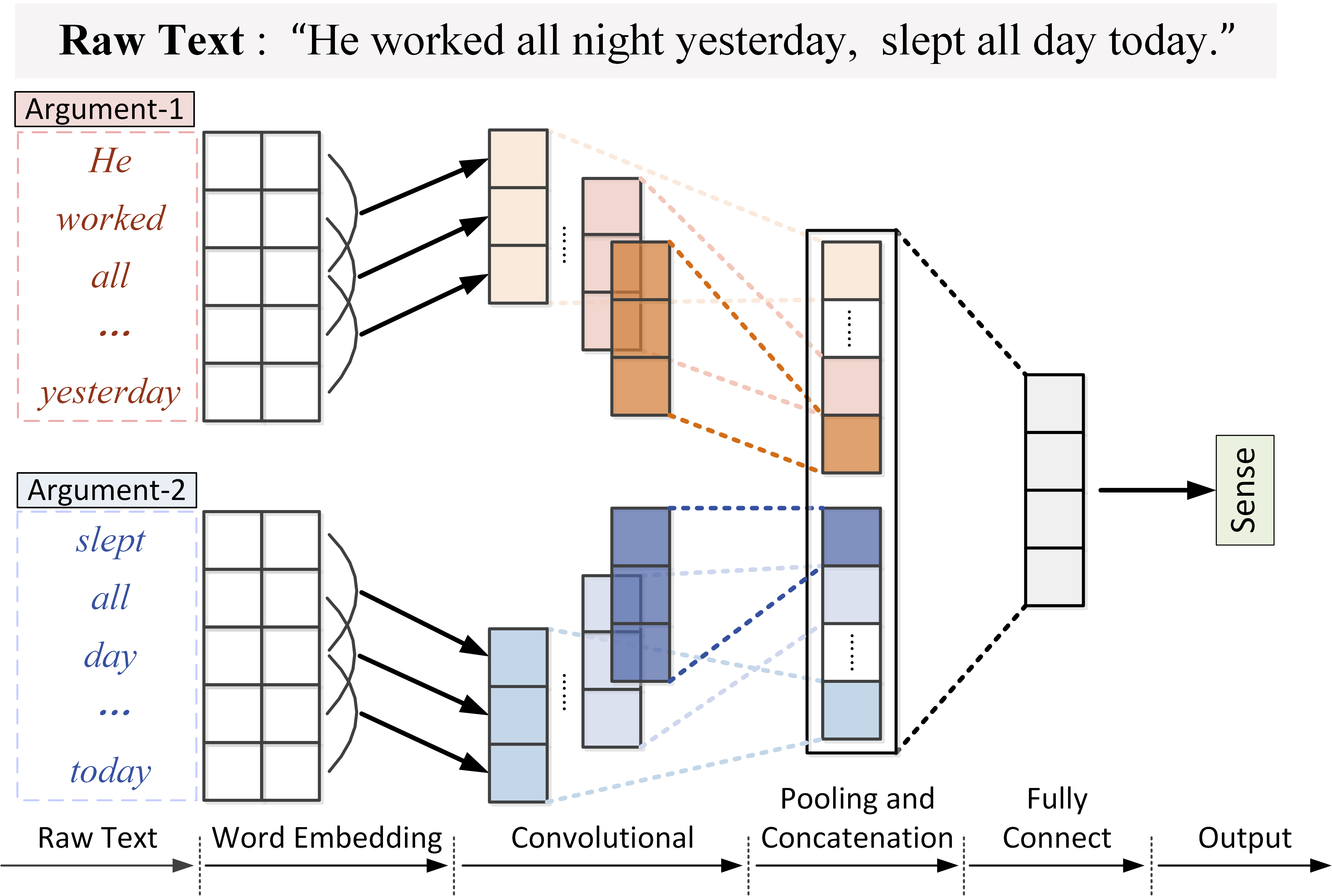}
	\caption{Illustration of using a convolution neural network for implicit discourse relation recognition.}
	\Description{A general structure of convolution neural network for implicit discourse relation recognition including raw text, word embedding layer, convolutinoal layer, pooling and concatenation layer, fully connect layer, and output.}
	\label{Fig:CNN}
\end{figure}

%卷积神经网络介绍及在语篇关系识别中的通用方法
For its great successes in many applications, the architecture and operation of a typical \textit{convolutional neural network} (CNN) have also been researched for the IDRR task~\cite{Zhang.B:et.al:2015:EMNLP, Qin.L:et.al:2016:EMNLP, Qin.L:et.al:2017:ACL, Nguyen.L.T:et.al:2019:ACL, Varia.S:et.al:2019:SIGDIAL, Liu.Y:et.al:2016:AAAI, Wu.C:et.al:2017:ACL, Wu.C:et.al:2019:NeuroComputing, She.X:et.al:2018:ACM, Li.H:et.al:2015:CCL}. A CNN generally consists of one or multiple convolution layers, pooling layers, fully connected layers, and a softmax output layer~\cite{Kim.Y:2014:EMNLP}, Fig.~\ref{Fig:CNN} illustrates the basic architecture of a CNN for the IDRR task. At the heart of a CNN is the convolutional operation, which is used to convert multiple consecutive input features in a sliding window into a single output feature. It is argued that such convolutional operations can help to capture semantic information in an argument.

\par
% CNN语篇关系识别的代表方法
The Shallow Convolutional Neural Network (SCNN)~\cite{Zhang.B:et.al:2015:EMNLP} might be the first CNN model for the IDRR task, which first obtains each argument representation via a single convolution layer and then concatenates a pair of argument representations to input a softmax layer for relation classification. At first, the word embeddings of an argument form a so-called argument matrix. The convolution operation is used to obtain the \textit{average}, \textit{min} and \textit{max} feature from each column of the argument matrix, by which an argument is converted into a three dimensional vector.

\par
% 门控+CNN语篇关系识别的方法
%Recently, the gated mechanism has gained popularity in neural networks, which is first introduced in the cells of recurrent neural networks.
%Here, the gates are mainly used for routing information from arguments, and the most crucial features are selected to capture the interactions between arguments.
%\xwmod{Recently, the gated mechanism has gained popularity in neural networks, that serves as a tool to route information in network training.}
Qin et al.~\cite{Qin.L:et.al:2016:EMNLP} proposed a stacking gated neural network, which can be divided into three parts: (1) two CNNs for argument representation learning; (2) a collaborative gated neural network for feature transformation; and (3) a softmax layer for relation classification. In their model, two gates work collaboratively to control the information flow of inner cells sequentially, resembling the logical AND operation in a probabilistic version.
%In addition, they fixed the lenghts of both arguments to be equivalent by truncating or zero-padding manner, and the two convolution operation share their parameters.

\par
% 隐式连接词+CNN语篇关系识别的方法
As mentioned in Section ~\ref{Sec:Implicit Connectives features}, a few ML methods have attempted to make use of annotated implicit connectives in the PDTB corpus~\cite{Zhou.Z:et.al:2010:COLING, Zhou.Z:et.al:2010:SIGDIAL}. However, these methods operates in a pipeline way, which might suffer from error propagations. Some researchers have also tried to include implicit connectives into CNN models~\cite{Qin.L:et.al:2017:ACL, Wu.C:et.al:2017:ACL, Wu.C:et.al:2019:NeuroComputing, Nguyen.L.T:et.al:2019:ACL}. For example, Wu et al.~\cite{Wu.C:et.al:2017:ACL, Wu.C:et.al:2019:NeuroComputing} proposed to learn discourse-specific word embeddings by performing connective classification on massive explicit discourse corpus, such that various discourse clues can be encoded into the word embeddings. Nguyen et al.~\cite{Nguyen.L.T:et.al:2019:ACL} employed a multi-task learning framework to predict implicit connectives and discourse relations simultaneously. Each predicted implicit connective corresponds to a kind of discourse relation, which can be used to facilitate knowledge transfer between two prediction tasks. Specifically, they proposed to project implicit connectives and discourse relations into a same latent representation space, and the knowledge from connective prediction are transferred to the relation prediction via a connective-relation mapping.

\par
%  对抗神经网络+CNN语篇关系识别的方法
Qin et al~\cite{Qin.L:et.al:2017:ACL} proposed to incorporate implicit connectives into an adversarial framework, which enables a self-calibrated imitation mechanism. The adversarial mechanism is an emerging method recently proposed for image generation~\cite{Goodfellow.I:et.al:2014:NIPS} and domain adaptation~\cite{Ganin.Y:et.al:2016:JourOfMachineLearnReach}, which learns to produce realistic samples through competition between a generator and a real/fake discriminator. In~\cite{Qin.L:et.al:2017:ACL}, the adversarial framework contains three main components: (1) a CNN for implicit relation recognition over raw arguments without access to connectives; (2) a connective-augmented CNN whose inputs are augmented with implicit connectives, serving as a feature emulation model; and (3) a discriminator to distinguish the features output from the two networks. The discriminator is a binary classifier modeled by a \textit{multi-layer perceptron} (MLP) and enhanced with a gated mechanism to identify the correct source of input features.

%%%%%%%%%%%%%%%%%%%%%%%%%%%%%%%%%%%%%%%%%%%%%%%%%%%%%%%%%%%%%%%%%%%%%%%%%%%%%%%%%%%%%
%% Subsec: Recurrent Neural Networks
%%%%%%%%%%%%%%%%%%%%%%%%%%%%%%%%%%%%%%%%%%%%%%%%%%%%%%%%%%%%%%%%%%%%%%%%%%%%%%%%%%%%%
\subsection{Recurrent Neural Networks}
% CNN的不足
The convolution operation in CNN models is executed on consecutive neighboring words, which can help capturing local contextual information for a word, but might ignore the word order information of an argument. As such, it cannot well capture potential inter-dependencies and interactions between distant words. In language modeling, an argument may be better treated as a sequence of words, such that not only word order information but also potential inter-dependencies in between distance words can be well exploited for representation learning.

\par
\begin{figure}[h]
	\centering
	\includegraphics[width=0.8\linewidth]{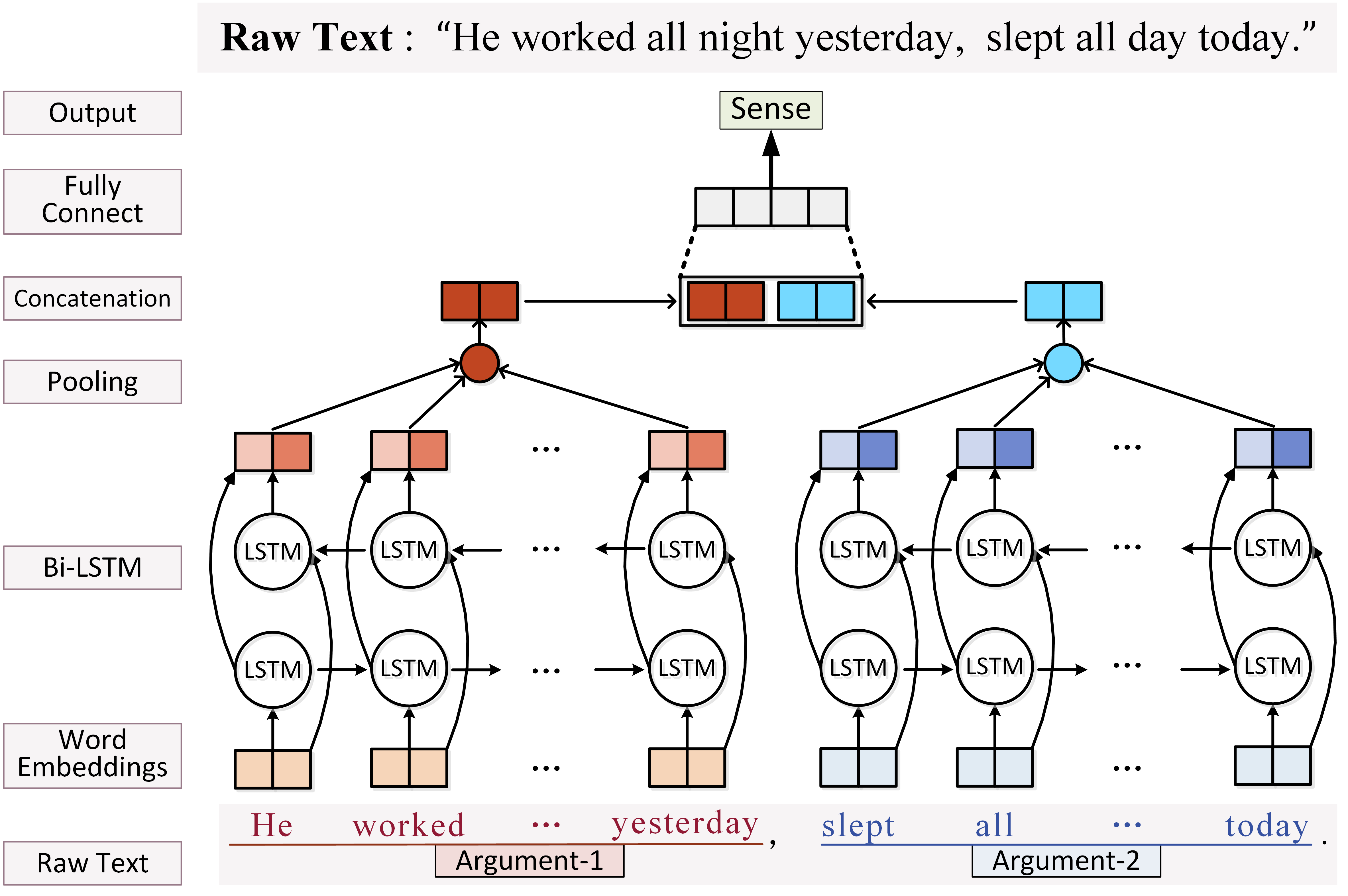}
	\caption{Illustration of using a recurrent neural network for implicit discourse relation recognition.}
	\Description{A general structure of recurrent neural network for implicit discourse relation recognition including raw text, word embedding layer, Bi-LSTM layer, pooling, layer, concatenation layer, fully connect layer, and output.}
	\label{Fig:RNN}
\end{figure}

A \textit{Recurrent Neural Network} (RNN) consists of a series of connected neurons to accept sequential data input and enables capturing some hidden information of the current word by taking consideration of all of its preceding words. Fig.~\ref{Fig:RNN} illustrates a basic RNN structure, in which a neuron computes the hidden state and output state for an input word and the hidden state of its preceding neuron also serves as its input. Such a recursive structure can effectively make use of the word sequence information of an argument.

%RNN语篇关系识别的代表方法
\par
Many RNN models have been proposed for the IDRR task~\cite{Rutherford.A:et.al:2017:EACL, Kishimoto.Y:et.al:2018:COLING, Yue.X:et.al:2018:Springer, Wu.C:et.al:2020:Elsevier}. For example, Rutherford et al.~\cite{Rutherford.A:et.al:2017:EACL} employed a \textit{Long Short-Term Memory} (LSTM) network, which extends memory cells with \textit{Gated Recurrent Units} (GRU). Specifically, each word of an argument is fed into a LSTM network, then a pooling function is applied over all words' output to produce an argument representation. The interaction between two arguments is modeled by multiple hidden layers that take a pair of argument representations as input and use an activation function and a softmax layer for relation classification. Yue et al.~\cite{Yue.X:et.al:2018:Springer} designed two RNN variants: One is an Externally Controllable LSTM network whose internal gates are controlled by externally supplied vector; The other is an Attention-Augmented GRU network which uses attention mechanism to augment the traditional GRU network. The two networks are stacked to incorporate arguments' interaction via a repeated reading strategy for relation classification.
%which equipped with externally controllable cells, which can be stacked to incorporate arguments' interactions into their representing process.
%Kishimoto et al.~\cite{Kishimoto.Y:et.al:2018:COLING} adopted a RNN which augments the input text with external knowledge and context.

\par
The syntactic tree of a sentence can be used to augment the basic RNN structure to encode syntactic dependencies in between words for argument representation learning~\cite{Rutherford.A:et.al:2017:EACL, Ji.Y:Eisenstein.J:2015:Trans.OfACL, Wang.Y:et.al:2017:IJCNLP, Otsuka.A:et.al:2015:PACLIC, Geng.R:et.al:2017:IEEE}. For example, Rutherford et al.~\cite{Rutherford.A:et.al:2017:EACL} proposed a tree-structured LSTM network built upon the syntactic tree of an argument, in which a neuron accepts not only an input word but also the hidden state from its syntactically dependent words. Ji and Eisenstein~\cite{Ji.Y:Eisenstein.J:2015:Trans.OfACL} proposed to further augment a tree-structured RNN with external entity mentions and some other linguistically informed features, like word-pair feature, to enrich input words' representations.

\par
% 段落或者篇章级别RNN语篇关系识别的方法
Generally, the semantics of a sentence cannot be interpreted independently from the rest of a paragraph or discourse. The aforementioned RNN models are not well in capturing long-range dependencies in a document. Some RNN-based models have been proposed to exploit inter-sentence contextual information~\cite{Ji.Y:et.al:2016:NAACL, Dai.Z:Huang.R:2018:NAACL}. For example, Ji et al.~\cite{Ji.Y:et.al:2016:NAACL} proposed a latent variable RNN, which encodes inter-sentence context information by a Document Context Language Model (DCLM)~\cite{Ji.Y:et.al:2015:arXiv}.
The DCLM encodes contextual information with two vectors: one representing the intra-sentence word-level context, and another representing inter-sentence context from the first sentence of document. These two vectors are linearly combined to generate a latent variable, which is used for each word of next sentence as the hidden state of its previous sentences. Dai and Huang~\cite{Dai.Z:Huang.R:2018:NAACL} proposed a paragraph-level RNN model that takes a sequence of discourse units as input and predicts a sequence of implicit discourse relations in a paragraph. A discourse unit is represented by the combination of word-level representations and argument-level representations, each produced by a \textit{bi-directional LSTM}  (BiLSTM) network in a paragraph.

%%%%%%%%%%%%%%%%%%%%%%%%%%%%%%%%%%%%%%%%%%%%%%%%%%%%%%%%%%%%%%%%%%%%%%%%%%%%%%%%%%%%%
%% Subsec: Hybrid Neural Network Models
%%%%%%%%%%%%%%%%%%%%%%%%%%%%%%%%%%%%%%%%%%%%%%%%%%%%%%%%%%%%%%%%%%%%%%%%%%%%%%%%%%%%%
\subsection{Hybrid Neural Network Models}
%CNN和RNN的不足
The aforementioned CNN and RNN models each has its own merits and demerits when used to capture semantic information of an argument and interaction between arguments for the IDRR task. A CNN can well encode local contextual information, but might lose the word order information. A RNN considers the word order information, however, cannot well capture the long-range dependencies of a paragraph or document. Some researchers have proposed various hybrid neural networks to enjoy the excellence of both models.

\par
%混合神经网络的代表方法
A common approach of building a hybrid neural model is to combine a CNN with a RNN~\cite{Guo.F:et.al:2019:ACCESS, Sun.Y:et.al:2019:NLPCC}. For example, Guo et al.~\cite{Guo.F:et.al:2019:ACCESS} proposed a Dynamic Chunk-based Max Pooling BiLSTM-CNN framework. Specifically, they first exploited a BiLSTM network to obtain semantic representation of an argument. They next adopted a convolutional layer to extract multi-granularity (n-gram) features from argument semantic representation. A dynamic chunk-based max-pooling strategy is followed up to divide each argument into several chunks according to the argument length. Finally, a fully connected layer with a softmax function is applied for relation classification. Sun et al.~\cite{Sun.Y:et.al:2019:NLPCC} proposed a multi-grain representation learning method, which concatenates word-level representation and larger-grain (such as phrase, chunk) representation of each argument. The word-level representation is obtained by a BiLSTM and the larger-grain representation is obtained by a few of convolution filters.

\par
%字级别的混合神经网络方法
Conventional word embedding models learn representation vectors at word level and ignore the information from characters that make up the word.
Character-aware model can learn word representations from its character compositions through a neural network that inputs the concatenation of character embeddings and outputs the representation of a word~\cite{Kim.Y:et.al:2016:AAAI, Ling.W:et.al:2015:EMNLP}.

\par
Some researchers proposed to combine character embeddings and word embeddings in neural networks~\cite{Qin.L:et.al:2016:COLING, Liu.X:et.al:2020:IJCAI}. For example, Qin et al.~\cite{Qin.L:et.al:2016:COLING} proposed a context-aware character-enhanced embedding model, which covers semantic information of three levels, that is, character, word, and sentence. Specifically, the model includes a character-level module and a word-level module. In the character-level module, the convolutional operation and max-pooling operation perform character-based word encoding and feature selection to obtain a sequence of character-based word representations. The sequence is transformed to a new sequence representation via a BiLSTM as the output of the character-level module. In the word-level module, the character-based word representations are concatenated into word embeddings. Finally, a CNN is used again to obtain sentence-level representations for an argument-pair, and followed by a conventional hidden layer and a softmax layer for the final classification.

\par
% 其他一些混合的神经网络
Some other hybrid neural networks have been designed through combining various models such as simplified topic model, gated convolutional network, factored tensor network, graph convolutional network, directed graphic model, sequence to sequence model, etc.~\cite{Zhang.B:et.al:2016:EMNLP, Shi.W:Demberg.V:2019:IWCS, Xu.S:et.al:2019:ACL, Zhang.Y:et.al:2021:NAACL}.
For example, Xu et al.~\cite{Xu.S:et.al:2019:ACL} proposed a Topic Tensor Network (TTN) model which combines a simplified topic model, a gated convolutional network and a factored tensor network. The simplified topic model can be interpreted as a neural network encoder that compresses the bag-of-word representation of arguments into a continuous hidden vector, i.e. the latent topic distribution, to provide topic-level representation. The gated convolutional network encodes each argument by stacking multiple gated convolutional layers to obtain sentence-level representation. They further proposed a factored tensor network to model both the sentence-level interactions and topic-level relevances using multi-slice tensors.

\par
Zhang et al.~\cite{Zhang.Y:et.al:2021:NAACL} proposed a Semantic Graph Convolutional Network (SGCN) to enhance the inter-argument semantic interaction. They first encoded each argument into a representation vector by a BiLSTM network. A semantic interaction graph is next built on an argument-pair, in which nodes are words from two arguments and an edge only exists in between one word in the first argument and another word in the second argument. The weight of an edge indicates the strength of semantic association between two connected nodes. A graph convolutional network is used to extract interactive features from the semantic interaction graph.

%%%%%%%%%%%%%%%%%%%%%%%%%%%%%%%%%%%%%%%%%%%%%%%%%%%%%%%%%%%%%%%%%%%%%%%%%%%%%%%%%%%%%
%% Subsec: Attention Mechanism
%%%%%%%%%%%%%%%%%%%%%%%%%%%%%%%%%%%%%%%%%%%%%%%%%%%%%%%%%%%%%%%%%%%%%%%%%%%%%%%%%%%%%
\subsection{Attention Mechanism}
%注意力机制介绍
Recently, various attention mechanisms have been proposed to augment representation learning for many NLP tasks~\cite{Wang.C:Wang.B:2020:WWW, Yang.Z:et.al:2019:NIPS, Zhou.P:et.al:2016:ACL}. Generally speaking, an attention mechanism is used to unequally treat each component of an input according to its importance to a given task~\cite{Vaswani.A:et.al:2017:NeurIPS}. The importance weights are computed based on an \textit{alignment model} that can be self-learned during the training process of a neural network. A weighted sum operation is normally executed to obtain a so-called \textit{context vector} as the new input.  For the IDRR task, it is also argued that different words in arguments contribute differently in learning argument representations and interactions.
% CNN+注意力机制的方法
\par
Some researchers have proposed to augment conventional neural models with attention mechanisms, such as the attention-based CNNs~\cite{Zhang.B:et.al:2018:Elsevier, Munir.K:et.al:2021:IEEE} and attention-based RNNs~\cite{Liu.Y:Li.S:2016:EMNLP, Fan.Z:et.al:2019:Comp.Sci., Lan.M:et.al:2017:EMNLP, Cai.D:Zhao.H:2017:IEA}. For example, Zhang et al.~\cite{Zhang.B:et.al:2018:Elsevier} proposed a full attention model that combines an \textit{inner attention model} to process internal argument information and an \textit{outer attention model} to exploit external world knowledge. In the inner attention model, they employed the SCNN~\cite{Zhang.B:et.al:2015:EMNLP} to encode each argument as its original representation, which is then used to compute the attention weight of each word by a score function. They further explored a cross-argument attention strategy that uses the original representation of one argument to obtain the attention attribution of the other one. In the outer attention model, they used an external semantic memory, i.e. a pre-trained word embedding matrix, which is considered to have encoded some world knowledge.

% RNN+注意力机制的方法
\par
Lan et al.~\cite{Lan.M:et.al:2017:EMNLP} presented a multi-task attention-based RNN to conduct argument representation learning. Fan et al.~\cite{Fan.Z:et.al:2019:Comp.Sci.} proposed a BiLSTM-based model combining self-attention mechanism and syntactic information. Liu and Li~\cite{Liu.Y:Li.S:2016:EMNLP} proposed a Neural Network with Multi-level Attention (NNMA) model, which includes an attention mechanism and external memories into a RNN model. Specifically, they first captured the general representation of each argument based on a BiLSTM model. Several attention levels are stacked upon the BiLSTM model and generate different weight vectors over an argument-pair, indicating what degree each word should be concerned. In each attention level, an external short-term memory is designed to store the information exploited in the previous levels and helps updating the argument representation.

\par
%一些基于注意力机制设计的新模型方法
%In addition, some novel neural models have been proposed that integrate attention mechanism for implicit discourse relation recognition~\cite{Bai.H:Zhao.H:2018:COLING, Guo.F:et.al:2018:COLING, Guo.F:et.al:2020:AAAI}.
Some other neural models also adopt attention mechanisms for the IDRR task~\cite{Bai.H:Zhao.H:2018:COLING, Guo.F:et.al:2018:COLING, Guo.F:et.al:2020:AAAI, Ronnqvist.S:et.al:2017:ACL, Liu.Y:et.al:2017:CCL, Ruan.H:et.al:2020:COLING}. For example, Bai and Zhao~\cite{Bai.H:Zhao.H:2018:COLING} proposed to learn different granularity levels of representations via an attention-augment neural model. First, the token sequences (i.e. characters, sub-words, and words) of an argument are firstly processed by several stacked encoder blocks (CNN or RNN) that are then processed by a bi-attention module in the argument-pair level for relation classification. Guo et al.~\cite{Guo.F:et.al:2018:COLING} proposed to use an interactive attention mechanism to enhance argument representation. They encoded each argument with a BiLSTM and computed the semantic connections in between all the word-pairs in two arguments as a pair-wise matrix. They next conducted column-wise and raw-wise softmax function on this matrix to get an interactive attention matrix. They further proposed a Knowledge-Enhanced Attentive Neural Network (KANN) framework~\cite{Guo.F:et.al:2020:AAAI} to integrate external knowledge via mapping a knowledge matrix into the interactive attention matrix.

\par
Recently, the \textit{Bidirectional Encoder Representation from Transformers }(BERT)~\cite{Devlin.J:et.al:2019:ACL} technique has been proposed as a language representation model, which is a recent breakthrough in the NLP field for its excellent performances in diverse tasks. The training model of BERT is a sequence-to-sequence transformer-based neural network architecture, which is solely based on attention mechanisms~\cite{Vaswani.A:et.al:2017:NeurIPS}. The BERT training procedure consists of two stages: pre-training and fine-tuning. In the pre-training, the BERT is trained by two unsupervised prediction tasks on large corpora like Wikipedia and BooksCorpus to capture contextual information, and output pre-trained words' representations. In the fine-tuning, the task-specific neural models can be designed upon the BERT with one additional output layer.

\par
Some researchers have proposed to exploit the BERT model for the IDRR task~\cite{Shi.W:Demberg.V:2019:EMNLP, Nie.A:et.al:2019:ACL, Kishimoto.Y:et.al:2020:LREC, Liu.X:et.al:2020:IJCAI, Jiang.D:He.J:2020:ACCESS, Li.X:et.al:2020:COLING, Zhou.M:et.al:2020:IEEE, Jiang.F:et.al:2021:IEEE}. In the pre-training, two arguments are concatenated into a single sequence as input, with a special token inserted at the beginning. In the fine-tuning, a classifier accepts the token sequence and outputs a discourse relation. Shi et al.~\cite{Shi.W:Demberg.V:2019:EMNLP} performed additional pre-training on domain-specific text. Nie et al.~\cite{Nie.A:et.al:2019:ACL} inserted an explicit connective prediction task in the pre-training.	Kishimoto et al.~\cite{Kishimoto.Y:et.al:2020:LREC} trained the BERT to jointly predict implicit connectives and discourse relations in fine-tuning steps as a multi-task learning task. Li et al.~\cite{Li.X:et.al:2020:COLING} used a penalty-based loss re-estimation method in classifier to strengthen the attention mechanism. Zhou et al.~\cite{Zhou.M:et.al:2020:IEEE} leveraged document-level discourse context to improve argument representation.

%%%%%%%%%%%%%%%%%%%%%%%%%%%%%%%%%%%%%%%%%%%%%%%%%%%%%%%%%%%%%%%%%%%%%%%%%%%%%%%%%%%%%
%% Subsec: Pair Interaction
%%%%%%%%%%%%%%%%%%%%%%%%%%%%%%%%%%%%%%%%%%%%%%%%%%%%%%%%%%%%%%%%%%%%%%%%%%%%%%%%%%%%%
\subsection{Neural Learning of Argument Pair Interaction}
\begin{figure}[h]
	\centering
	\includegraphics[width=\linewidth]{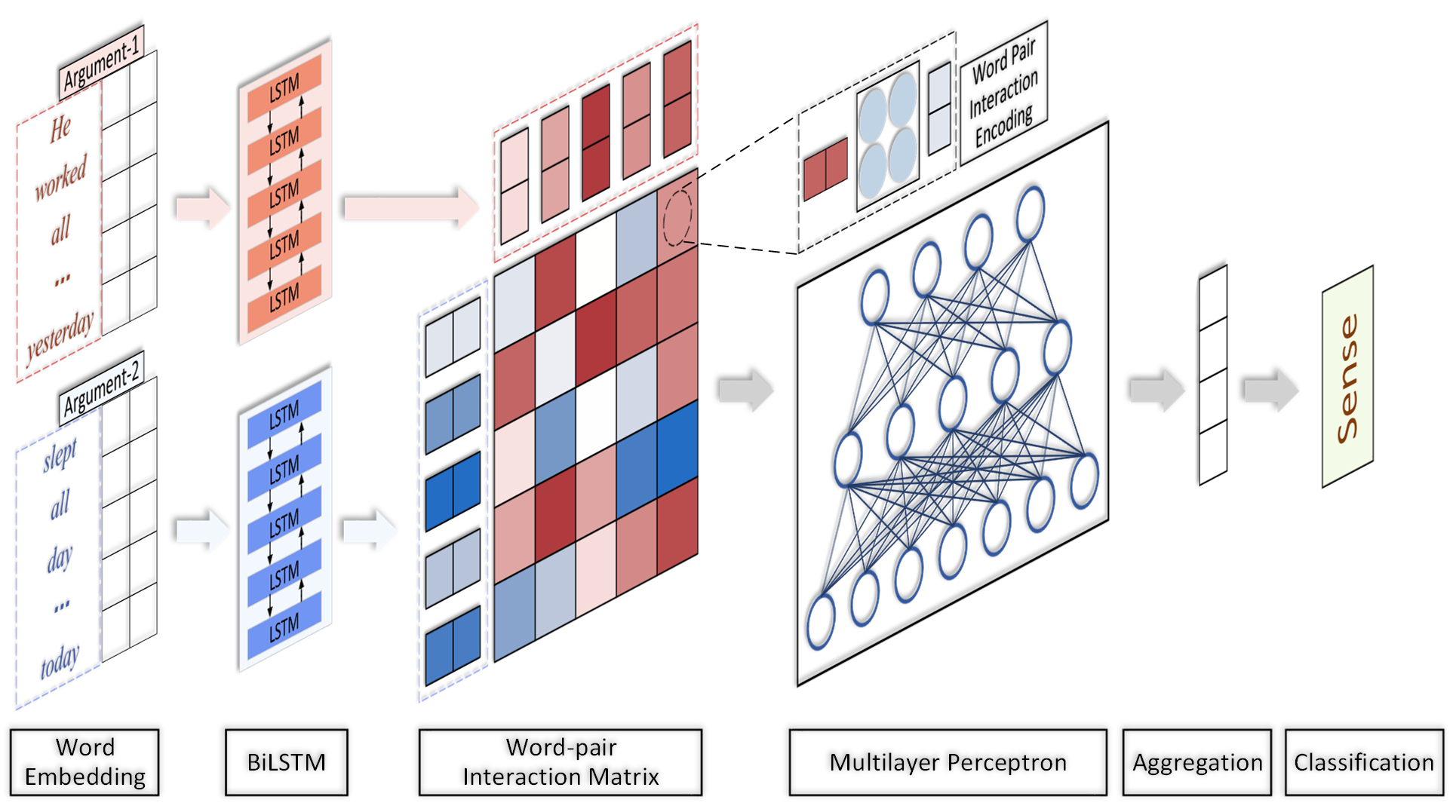}
	\caption{Illustration of using a word-pair interaction matrix for implicit discourse relation recognition.}
	\Description{A word-pair interaction matrix generated by the word-pairs in terms of the cartesian product of two arguments, that be transformed into an argument-pair representation for relation classification.}
	\label{Fig:Interaction}
\end{figure}

% 事件句子对中的词语对，在机器学习方法中被证明是有效的特征。也可以用于深度学习中。一方面用词嵌入表示词语特征，另一方面用神经网络作为分类器。
A word-pair in the IDRR task refers to a pair of two words, one from the first argument and another from the second argument. It has been experimented that such word-pairs can help improving the IDRR performance in those conventional ML methods (see Section ~\ref{Sec:Machine Learning}). Some researchers proposed to use word embedding based features to replace hand-crafted word-pair features in order to capture interactions between arguments via neural networks. As illustrated in Fig.~\ref{Fig:Interaction}, words in each argument can be firstly encoded via a BiLSTM network. Then a \textit{word-pair interaction encoding} can be computed as the Cartesian product of the hidden states of two words each from one argument. The \textit{word-pair interaction matrix} is next transformed into an argument-pair representation via a neural network for relation classification~\cite{Chen.J:et.al:2016:AAAI, Chen.J:et.al:2016:ACL}.

%词语间的互动
\par
Chen et al.~\cite{Chen.J:et.al:2016:AAAI} introduced a mixed generative-discriminative framework to word embedding offsets as the elements of the word-pair interaction matrix. As the length of each argument is different, they further used a Fisher Kernel~\cite{Jaakkola.T:Haussler.D:1999:NeurIPS} to aggregate the word-pair interaction matrix into a fixed length argument-pair representation for relation classification. Chen et al.~\cite{Chen.J:et.al:2016:ACL} proposed to construct a relevance score word-pair interaction matrix for an argument-pair, in which the relevance score of each word-pair is computed by a bilinear model~\cite{Sutskever.I:et.al:2008:NeurIPS, Jenatton.R:et.al:2012:NeurIPS} and a single layer neural model~\cite{Collobert.R:Weston.J:2008:ICML}. Varia et al.~\cite{Varia.S:et.al:2019:SIGDIAL} constructed a word-pair interaction matrix by using the concatenation of the words' embeddings of a word-pair as a matrix element. A CNN is then used to learn an argument-pair representation from the word-pair interaction matrix.

%整个事件句子之间的互动
\par
Some researchers utilized the word-pair interaction matrix to represent an argument as a probability distribution vector in order to capture some contextual interactions between two arguments. For example, Lei et al.~\cite{Lei.W:et.al:2017:IJCAI} proposed a Simple Word Interaction Model (SWIM) to learn an argument representation from word-pair matrix. They computed linear and quadratic relations in word-pair embeddings to form an word-pair interaction score matrix. The SWIM then computes each argument a representation as the concatenation of word-pair embeddings, and weights them by the word-pair interaction score matrix to account its importance.

\par
%参考TransE（一种实体关系识别方法），计算相对位置的方法。
Motivated by the Translating Embeddings Model (TransE) for entity relation extraction~\cite{Bordes.A:et.al:2013:NIPS}, some work intended to embed the first argument into one latent space and the second argument into another latent space, so as to model their relation by the direction and position between two arguments' representations in the two latent spaces~\cite{Li.H:et.al:2016:NLPCC, Li.H:et.al:2017:ACM, He.R:et.al:2020:ACL}. Intuitively, the same sense of a discourse relation may have similar direction and position information in the two latent spaces, by which new discourse relations can be recognized by using the translating operation in those Trans-series models~\cite{Xiao.H:et.al:2016:ACL, Wang.Z:et.al:2014:AAAI, Ji.G:et.al:2015:ACL, Lin.Y:et.al:2015:AAAI}. For example, He et al.~\cite{He.R:et.al:2020:ACL} proposed a novel TransS-driven joint learning architecture, which translates a discourse relation into a low-dimensional latent space to mine some latent geometric structure information of arguments for relation classification. Li et al.~\cite{Li.H:et.al:2016:NLPCC, Li.H:et.al:2017:ACM} designed a multi-view Tensor and Trans Neural Network (TTNN) model, in which the Tensor module focuses on the interactions between arguments and the Trans module explores the direction and position information between arguments in two latent spaces.

%%%%%%%%%%%%%%%%%%%%%%%%%%%%%%%%%%%%%%%%%%%%%%%%%%%%%%%%%%%%%%%%%%%%%%%%%%%%%%%%%%%%%
%%
%% Sec: Expanding Train data
%%
%%%%%%%%%%%%%%%%%%%%%%%%%%%%%%%%%%%%%%%%%%%%%%%%%%%%%%%%%%%%%%%%%%%%%%%%%%%%%%%%%%%%%
%\section{Implicit discourse relation recognition based on semi-supervised learning}\label{Sec:Add Data}
\section{Data Expansion based on Semi-supervised Learning}\label{Sec:Add Data}
%监督学习需要大量语料
The aforementioned fully-supervised ML and DL methods require a large amount of labeled data for model training. However, the available annotated corpus is limited. Manually annotating data is a time-consuming process which also requires domain expertise and professional knowledge in most cases. Semi-supervised learning is a combination of supervised and unsupervised learning, that uses a small set of labeled data to generate more training data from unlabeled data for model training. In this section, we review such solutions in the literature, mainly focusing on how they expand labeled data and how classification models can be trained from mixed data.

%%%%%%%%%%%%%%%%%%%%%%%%%%%%%%%%%%%%%%%%%%%%%%%%%%%%%%%%%%%%%%%%%%%%%%%%%%%%%%%%%%%%%
%% Subsec: Data Expansion from Explicit Discourse Relations
%%%%%%%%%%%%%%%%%%%%%%%%%%%%%%%%%%%%%%%%%%%%%%%%%%%%%%%%%%%%%%%%%%%%%%%%%%%%%%%%%%%%%
\subsection{Data Expansion from Explicit Discourse Relations}
%显式连接词扩展语料的优势和不足：直接删除显式连接词，可以被看做是隐式关系，但实际运用（模型训练）中效果不好，会由噪音
A straightforward solution is to exploit an \textit{explicit relation corpus} (ERC) to expand an \textit{implicit relation corpus} (IRC). The difference between the two types of corpus is that the connectives in an ERC are from raw texts, but in an IRC the connectives are manually inserted by annotators as a part of label for model training. Early methods obtained an expanded IRC via directly removing connectives in an ERC~\cite{Marcu.D:Echihabi.A:2002:ACL, Sporleder.C:Lascarides.A:2008:NaturalLanguageEng.}. Obviously, such approaches are easy to implement to obtain a so-called \textit{synthetic} IRC. However, a synthetic IRC may exhibit much differences from the native IRC. Experiment results have shown that training on a synthetic IRC is not necessarily a good strategy due to the linguistically dissimilarity between explicit and implicit samples~\cite{Sporleder.C:Lascarides.A:2008:NaturalLanguageEng.}.

%通过省略率和上下文差异来挑选好的例子，进行语料扩展
\par
%To avoid including noisy samples,
Researchers have proposed various approaches to select the most suitable \textit{explicit samples} in ERC into a synthetic IRC~\cite{Rutherford.A:Xue.N:2015:NAACL, Xu.Y:et.al:2018:EMNLP, Wu.C:et.al:2017:Springer}.
For example, Rutherford and Xue~\cite{Rutherford.A:Xue.N:2015:NAACL} proposed two selection criteria: \textit{omission rate} and \textit{context differential}. They hypothesized that those connectives often omitted or insensitive to semantic contexts are good candidate explicit samples.
The omission rate of a connective is computed by the ratio of its occurrences in an IRC to its total occurrences in both ERC and IRC. They used a Jensen-Shannon Divergence (JSD) to measure the context differential between the context of a connective in implicit and explicit discourse relation. If a connective has a high omission rate or a low context differential, it can be included into a syntactic IRC.
%Xu et al.~\cite{Xu.Y:et.al:2018:EMNLP} carried out an experiment of sampling, in which a simple active learning approach is used, so as to take the informative instances.
%Wu et al.~\cite{Wu.C:et.al:2017:Springer} proposed to use co-training model~\cite{Blum.A:Mitchell.T:1998:ACM} to select useful instances from artificial implicit data.

%混合天然隐式关系和人工创造的隐式关系（删除显式中的连接词）来挑选好的数据集
\par
Wang et al.~\cite{Wang.X:et.al:2012:COLING} considered that an effective sample for training should have distinct characteristics to signify some discourse relation, called a \textit{typical sample}, which can be either from a native IRC or a synthetic IRC. They proposed a Single Centroid Clustering (SCC) to select typical samples as training data. Specifically, the SCC computes a centroid based on the most significant features for each discourse relation and then uses the relation centroid to reassign samples as either typical or atypical. Ji et al.~\cite{Ji.Y:et.al:2015:EMNLP} argued why explicit and implicit samples may be linguistically dissimilar is due to domain mismatch. They proposed to construct a synthetic IRC by using two domain adaptation methods: (1) feature representation learning: mapping the source domain (explicit) and target domain (implicit) to a shared latent feature space; (2) re-sampling: modifying the distribution of relation sense in explicit to match the distribution over implicit relations.

%%%%%%%%%%%%%%%%%%%%%%%%%%%%%%%%%%%%%%%%%%%%%%%%%%%%%%%%%%%%%%%%%%%%%%%%%%%%%%%%%%%%%
%% Subsec: Data Expansion from Multi-Language Data
%%%%%%%%%%%%%%%%%%%%%%%%%%%%%%%%%%%%%%%%%%%%%%%%%%%%%%%%%%%%%%%%%%%%%%%%%%%%%%%%%%%%%
\subsection{Data Expansion from Multi-Language Data}
%一种语言的隐式关系在另一种语言中可能是显式关系，可以直接在 “对齐”（语篇关系在不同的语言中是一一对应的）的不同语言语篇关系语料，进行语料扩展
A same argument-pair and their relation may be described in different languages; While the labeled data from one language is highly possible to convey similar information in another language. Motivated from such considerations, some approaches have been proposed to utilize multi-language corpora~\cite{Wu.C:et.al:2016:EMNLP, Zhou.Y:Xue.N:2012:ACL, Wu.C:et.al:2017:NeuroComputing, Shi.W:et.al:2017:IJCNLP, Lu.Y:et.al:2018:Springer}. For example, Wu et al.~\cite{Wu.C:et.al:2016:EMNLP, Wu.C:et.al:2017:NeuroComputing} proposed to construct a bilingually-constrained synthetic IRC, which includes an implicit sample in one language if its corresponding one in another language is an explicit sample. Zhou and Xue~\cite{Zhou.Y:Xue.N:2012:ACL} showed that the connectives in Chinese omit much more frequently than those in English.
They constructed their bilingually-constrained synthetic IRC from a Chinese-English sentence-aligned corpus according to the mismatch of connectives in the two languages.
%bilingual implicit or explicit mismatch.

%通过翻译系统，在翻译过程中有可能删除或者增加连接词，来扩展语料
\par
People sometimes omit connectives during translation, or insert connectives not originally present in the source text.
Lu et al.~\cite{Lu.Y:et.al:2018:Springer} used machine translation to generate Chinese implicit samples from a labeled English discourse corpus. Shi et al.\cite{Shi.W:et.al:2017:IJCNLP} proposed to automatically extract implicit samples from parallel corpora via back-translation.
They used a sentence-aligned parallel corpus with English and French discourses. The potential implicit discourse relations in English are back-translated from French with explicit connectives. The inserted connective in English disambiguates the originally implicit discourse relation with a relation confidence score.

%%%%%%%%%%%%%%%%%%%%%%%%%%%%%%%%%%%%%%%%%%%%%%%%%%%%%%%%%%%%%%%%%%%%%%%%%%%%%%%%%%%%%
%% Subsec: Joint Data Expansion and Model Training
%%%%%%%%%%%%%%%%%%%%%%%%%%%%%%%%%%%%%%%%%%%%%%%%%%%%%%%%%%%%%%%%%%%%%%%%%%%%%%%%%%%%%
\subsection{Joint Data Expansion and Model Training}
%Boostrapping方法：通过少量语料训练模型，然后用在未标注文本上识别关系，取置信度高得加入训练集，迭代训练
Although the golden labeled data are few and the synthetic data may contain some linguistically dissimilarity, they can be iteratively used for model training~\cite{Fisher.R:Simmons.R:2015:IJCNLP, Zhou.M:et.al:2020:IJCNN, Hernault.H:et.al:2010:EMNLP, Nishida.N:Nakayama.H:2018:SIGDIAL, Lan.M:et.al:2013:ACL, Li.J:et.al:2014:COLING, Kurfali.M:Ostling.R:2019:SIGDIAL, Braud.C:Denis:P:2014:COING}. For example, Fisher et al.~\cite{Fisher.R:Simmons.R:2015:IJCNLP} and Zhou et al.~\cite{Zhou.M:et.al:2020:IJCNN} used the \textit{bootstrapping} method in an iterative training model. The basic idea is to train the model using golden labeled data first, and then the predictions on unlabeled data with high confidence are added as training data for iterative training.
%Hernault et al.~\cite{Hernault.H:et.al:2010:EMNLP} analysed the co-occurring features in abundant, freely-available unlabeled data, which is then taken into account for extending the feature vectors for a classifier.

\par
%多任务机器学习方法
The idea of multi-task learning method is to solve a main task together with other related auxiliary tasks at the same time. Lan et al.~\cite{Lan.M:et.al:2013:ACL} designed a multi-task learning model, in which the main task is based on the original IRC while the auxiliary task is based on a synthetic IRC. According to the principle of multi-task learning, their model can be optimized by the shared part of the main task and the auxiliary tasks. Liu et al.~\cite{Liu.Y:et.al:2016:AAAI} found that under different discourse annotation frameworks existing multiple corpora have some internal connections. They exploited such different corpora to design a CNN-based multi-task learning system to synthesize other tasks by learning both unique and shared representations for each task.

%跨语言得方法	
\par
Multi-lingual corpora can be utilized to train a joint IDRR model in different languages. Li et al.~\cite{Li.J:et.al:2014:COLING} used projections across a parallel corpus with Chinese and English for exploiting implicit samples. The main idea is to first recognize English implicit discourse relations, then project the predicted sense labels in English onto the corresponding Chinese samples. Kurfali and Ostling~\cite{Kurfali.M:Ostling.R:2019:SIGDIAL} represented arguments with multi-language sentence embedding via a pre-trained LASER model~\cite{Artetxe.M:Schwenk.H:2019:TACL}, and fed them into a feed forward network~\cite{Rutherford.A:et.al:2017:EACL}.
She et al.~\cite{She.X:et.al:2018:ACM} employed a distributed representation of hierarchical semantic components from different languages as classification triggers.
%%%%%%%%%%%%%%%%%%%%%%%%%%%%%%%%%%%%%%%%%%%%%%%%%%%%%%%%%%%%%%%%%%%%%%%%%%%%%%%%%%%%%
%%
%% Sec: Performance Comparison
%%
%%%%%%%%%%%%%%%%%%%%%%%%%%%%%%%%%%%%%%%%%%%%%%%%%%%%%%%%%%%%%%%%%%%%%%%%%%%%%%%%%%%%%
\section{Performance Comparison}\label{Sec:Performance}
%  In this section, we first introduce the partitioning of the PDTB dataset and evaluation metrics, and then compare those algorithms experimented on the PDTB dataset with the standard evaluation procedures.

%%%%%%%%%%%%%%%%%%%%%%%%%%%%%%%%%%%%%%%%%%%%%%%%%%%%%%%%%%%%%%%%%%%%%%%%%%%%%%%%%%%%%
%% Subsec: Dataset
%%%%%%%%%%%%%%%%%%%%%%%%%%%%%%%%%%%%%%%%%%%%%%%%%%%%%%%%%%%%%%%%%%%%%%%%%%%%%%%%%%%%%
%\subsection{Dataset setting}
%数据集划分
The PDTB corpus provides a hierarchical structure for discourse relation senses. But most of researches have mainly focused on the Level-1 relation senses, that is, the \textsf{Comparison}, \textsf{Contingency}, \textsf{Expansion} and \textsf{Temporal} relation sense, as those senses of Level-2 types and Level-3 subtypes are too fine-grained with very few training samples. For experimentation, the PDTB corpus is normally split into three parts, namely, a training set, a development set, and a test set. In order to make fair comparisons, most researches use the following dataset partition~\cite{Pitler.E:et.al:2009:ACL}: Section 0-2 and 21-22 of the PDTB dataset are used as the development set and the test set, respectively. Section 2-20 are randomly down-sampled to construct training sets each with both positive and negative samples with respect to a target relation. Table~\ref{Tab:Statistics} presents such an example of dataset division. We compare only those models using this division for the Level-1 relation recognition; While a few of other models use different dataset divisions are not included for comparison.

\begin{table*}
	\centering
	\caption{Statistics of positive and negative implicit relation instances in training, development and test sets for each relation sense on PDTB 2.0.}
	\begin{tabular}{c|ccc}
		\hline
		\textbf{Relation} & \textbf{Train } & \textbf{Dev.} & \textbf{Test} \\
		\hline
		\textbf{Comparison} & 1942/1942 & 197/986 & 152/894 \\
		\textbf{Contigency} & 3342/3342 & 295/888 & 297/767 \\
		\textbf{Expansion} & 7004/7004 & 671/512 & 574/472 \\
		\textbf{Temporal} & 760/760 & 64/1119 & 85/961 \\
		\hline
	\end{tabular}%
	\label{Tab:Statistics}%
\end{table*}%

%In addition, the nature of task and dataset pose at least two problems in creating a classifier. Except for the classification task requires a large number of features, some of which are too rare and inconducive to parameter estimation.
%The label of sense distribution is highly imbalanced and this might degrade the performance of the classifiers, as present in Table.~\ref{Tab:Statistics}.
%Some works usually discard the data in training set to balance the label distribution.
%For example, Park and Cardie(2012)~\cite{Park.J:Cardie.C:2012:SIGDIAL} and Wang et al. (2012)~\cite{Wang.X:et.al:2012:COLING} addressed these problems directly by optimally select a subset of features and training samples.
%Rutherford and Xue(2014)~\cite{Rutherford.A:Xue.N:2014:ACL} reweighted the training samples in each class during parameter estimation such that the performance on the development set is maximized.

%%%%%%%%%%%%%%%%%%%%%%%%%%%%%%%%%%%%%%%%%%%%%%%%%%%%%%%%%%%%%%%%%%%%%%%%%%%%%%%%%%%%%
%% Subsec: Metrics
%%%%%%%%%%%%%%%%%%%%%%%%%%%%%%%%%%%%%%%%%%%%%%%%%%%%%%%%%%%%%%%%%%%%%%%%%%%%%%%%%%%%%
%\subsection{Evaluation Metrics}
%由于各种关系的数据分布不均衡，因此用4个二分类
\par
As shown in Table~\ref{Tab:Statistics}, the samples of the Level-1 four discourse relations are imbalanced. For example, the samples of the \textsf{Expansion} relation occupy more than 50\%; While those of the \textsf{Temporal} relation constitute only 5\% of  the PDTB corpus. Given such an imbalanced dataset, the IDRR task has been often formulated as four \textit{one-against-all} binary classification, with the \textit{one} for each Level-1 relation sense and the \textit{all} for the other three relation senses. Furthermore, the positive and negative samples in the test set are also not balanced. For example, there are only 8.8\% positive samples in the test set of \textsf{Temporal} sense. For these considerations, most researches adopt the \textit{F1 score} as the main performance metric, which is a harmonic mean of the \textit{precision} and \textit{recall} performance metric.

\par
Table~\ref{Tab:ML_Performance} and Table~\ref{Tab:DL_Performance}, respectively, present the reported F1 score by those traditional machine learning and deep learning algorithms for the English IDRR task. It is worth of noting that some algorithms also depend on certain upstream tasks' results; While most of them have directly used the gold annotations in the PDTB corpus as a part of the input for the IDRR task. Table~\ref{Tab:ML_Performance} and Table~\ref{Tab:DL_Performance} show that in general advanced neural models can achieve better performance than traditional machine learning algorithms. The state-of-the-art performances of \textsf{Temporal}, \textsf{Comparison}, and \textsf{Contingency} relation classification have improved 46\%, 32\%, and 15\% respectively; While the \textsf{Expansion} relation has a 2\% slight improvement. This can be attributed to the powerful capabilities of neural networks for learning deep yet more comprehensive context-aware and/or syntactic-aware word and argument representations.

%
%However, we can observe that most deep learning algorithms have relatively low F1 scores in all the four relation recognitions. \wbnote{This may be attributed to ...} As presented in Table~\ref{Tab:ML_Performance}, the \textsf{Expansion} relation can achieve above 70\% and the \textsf{Contingency} relation can achieve 60\% F1 score; While the other two relations are often with lower than 50\% F1 scores. The four-way relation classification can achieve above 50\% F1 score.

\par
Furthermore, we can also observe from Table~\ref{Tab:DL_Performance} that attention mechanism has an effective improvement in the four-way relation classification and the \textsf{Comparison}, \textsf{Contingency}, and \textsf{Temporal} relation recognition. This indicates that attention mechanisms can enhance representation learning by unequally treating each input argument component according to its importance. Finally, we note that the model developed by Kishimoto et al. (2020)~\cite{Kishimoto.Y:et.al:2020:LREC} and Liu et al. (2020)~\cite{Liu.X:et.al:2020:IJCAI} that have applied the recent BERT model for the IDRR task can achieve over 70\% F1 score in all the four relation recognitions and over 60\% F1 score in four-way relation classification respectively. This indicates that the IDRR task can greatly benefit from the informative pre-trained language model trained from massive amounts of unlabeled text with diverse backgrounds.

\begin{table*}
	\centering
	\caption{Performance comparison of implicit discourse relation recognition based on machine learning on the PDTB 2.0 dataset.}
	\label{Tab:ML_Performance}
	\resizebox{0.94\columnwidth}{!}{
		\renewcommand\arraystretch{1}
		\begin{tabular}{c|c|c c c c}
			\hline
			\multirow{2}{*}{\textbf{System}}   & \multirow{2}{*}{\textbf{Classifier}} & \multicolumn{4}{c}{\textbf{Result (F1-score)}}  \\
			\cline{3-6}	
			&       & \textbf{Comp.} & \textbf{Cont.} & \textbf{Exp.} & \textbf{Temp.} \\
			\hline
			Pitler et al. (2009) ~\cite{Pitler.E:et.al:2009:ACL} & Naive Bayes & 21.96\% & 47.13\% & \underline{76.42\%} & 16.76\% \\
			Zhou et al. (2010) ~\cite{Zhou.Z:et.al:2010:COLING} & Support Vector Machine & 31.79\% & 47.16\% & 70.11\% & 20.30\% \\
			Park \& Cardie (2012) ~\cite{Park.J:Cardie.C:2012:SIGDIAL} &  Naive Bayes & 31.32\% & 49.82\% & \textbf{79.22\%} & 26.57\% \\
			Xu et al. (2012) ~\cite{Xu.Y:et.al:2012:IJCNN_IEEE} & Maximum Entropy & 24.45\% & 50.37\% & 63.44\% & 16.91\% \\
			Biran \& McKeown (2013) ~\cite{Biran.O:McKeown.K:2013:ACL} & Naive Bayes & 25.40\% & 46.94\% & 75.87\% & 20.23\% \\
			Rutherford \& Xue (2014) ~\cite{Rutherford.A:Xue.N:2014:ACL} &  Naive Bayes & 39.70\% & 54.42\% & 70.23\% & 28.69\% \\
			Li et al. (2015) ~\cite{Li.H:et.al:2015:CCL} &   Support Vector Machine & \underline{40.55\%} & 55.14\% & 70.71\% & \textbf{35.00\%} \\
			Li et al. (2016) ~\cite{Li.S:et.al:2016:Jour.OfChineseInfor.Proc.} &  Maximum Entropy & 34.24\% & 50.32\% & 66.98\% & 20.08\% \\	
			Roth (2017) ~\cite{Roth.M:2017:IWCS} & Logistic Regression & 37.00\% & \underline{56.30\%} & 69.40\% & \underline{32.10\%} \\
			Wei et al. (2018) ~\cite{Lei.W:et.al:2018:AAAI} &  Naive Bayes & \textbf{43.24\%} & \textbf{57.82\%} & 72.88\% & 29.10\% \\	
			\hline
			\multicolumn{2}{c|}{Average} & 32.97\% & 51.54\% & 71.53\% & 24.57\% \\
			\hline
	\end{tabular}}
\end{table*}

\begin{table*}
	\centering
	\caption{Performance comparison of implicit discourse relation recognition based on deep learning on the PDTB 2.0 dataset.}
	\label{Tab:DL_Performance}
	\resizebox{0.94\columnwidth}{!}{
		\renewcommand\arraystretch{0.9}
		\begin{tabular}{c|c c|c c c c}
			\hline
			\multirow{2}{*}{\textbf{System}}   & \multicolumn{2}{c|}{\textbf{Four-way Classifcation}} & \multicolumn{4}{c}{\textbf{Binary Classification (F1)}}  \\
			\cline{2-7}	
			&   \textbf{F1} & \textbf{Acc}  & \textbf{Comp.} & \textbf{Cont.} & \textbf{Exp.} & \textbf{Temp.} \\
			\hline
			%%%%%%%%%%%%%%%%%%%%%%%%% CNN
			\multicolumn{7}{c}{Convolutional neural networks} \\
			\hline			
			Zhang et al. (2015) ~\cite{Zhang.B:et.al:2015:EMNLP}     & -                   & -       & 33.22\%              & 52.04\% & 69.59\% & 30.54\% \\
			Liu et al. (2016) ~\cite{Liu.Y:et.al:2016:AAAI}          & 44.98\%             & 57.27\% & 34.65\%              & 46.09\% & 69.88\% & 31.82\% \\
			Qin et al. (2016) ~\cite{Qin.L:et.al:2016:EMNLP}         & -                   & -       & \underline{41.55\%}              & \textbf{57.32\%} & 71.50\% & 35.43\% \\
			Wu et al. (2017) ~\cite{Wu.C:et.al:2017:ACL}             & 44.84\%             & \textbf{58.85\%} &  35.10\%             & 47.82\% & 70.66\% & 25.81\% \\
			Qin et al. (2017) ~\cite{Qin.L:et.al:2017:ACL}           & -                   & -       &  40.87\%             & 54.56\% & 72.38\% & 36.20\% \\
			She et al. (2018) ~\cite{She.X:et.al:2018:ACM}           & 43.00\%             & 55.40\% &  35.90\%             & 52.50\% & \textbf{77.00\%} & 18.20\% \\
			Wu et al. (2019) ~\cite{Wu.C:et.al:2019:NeuroComputing}  & \underline{48.39\%} & \underline{58.36\%} &  33.18\%             & 48.99\% & 69.10\% & \textbf{42.31\%} \\
			Nguyen et al. (2019) ~\cite{Nguyen.L.T:et.al:2019:ACL}   & \textbf{53.00\%}    & -       &  \textbf{48.44\%} & \underline{56.84\%} & \underline{73.66\%} & \underline{38.60\%} \\
			\hline
			Average & 46.84\% & 57.47\% & 37.86\% & 52.02\% & 71.72\% & 32.36 \\
			\hline
			%%%%%%%%%%%%%%%%%%%%%%%%% RNN
			\multicolumn{7}{c}{Recurrent neural networks} \\
			\hline
			Ji \& Eisenstein (2015) ~\cite{Ji.Y:Eisenstein.J:2015:Trans.OfACL} & - & -  & 35.93\% & 52.78\% & \underline{80.02\%} & 27.63\% \\
			Geng et al. (2017) ~\cite{Geng.R:et.al:2017:IEEE} & \underline{44.20\%} & \textbf{62.40\%}  & 35.40\% & 53.80\% & \textbf{81.40\%} & 32.80\% \\
			Yue et al. (2018) ~\cite{Yue.X:et.al:2018:Springer} & - & -  & \underline{40.03\%} & \underline{56.38\%} & 70.10\% & \underline{32.85\%} \\
			Dai \& Huang (2018) ~\cite{Dai.Z:Huang.R:2018:NAACL} & \textbf{51.84\%} & \underline{59.75\%}  & \textbf{46.79\%} & \textbf{57.09\%} & 70.41\% & \textbf{45.61\%} \\
			\hline
			Average   & 48.02\% & 61.08\% & 39.54\% & 55.01\% & 75.48\% & 34.72\% \\
			\hline
			%%%%%%%%%%%%%%%%%%%%%%%%% Hybrid
			\multicolumn{7}{c}{Hybrid neural networks} \\
			\hline
			Zhang et al. (2016) ~\cite{Zhang.B:et.al:2016:EMNLP} & - & - & 35.88\% & 50.56\% & 71.48\% & 29.54\% \\
			Qin et al. (2016) ~\cite{Qin.L:et.al:2016:COLING}  & - & - & 38.67\% & 54.91\% & \textbf{80.66\%} & 32.76\% \\
			Guo et al. (2019) ~\cite{Guo.F:et.al:2019:ACCESS} & 46.39\% & \underline{57.69\%} & 39.84\% & \underline{56.31\%} & 71.59\% & 37.76\% \\
			Sun et al. (2019) ~\cite{Sun.Y:et.al:2019:NLPCC} & - & - & \underline{45.10\%} & 54.72\% & \underline{73.30\%} & \underline{40.18\%} \\
			Shi \& Demberg (2019) ~\cite{Shi.W:Demberg.V:2019:IWCS} & \underline{46.40\%} & \textbf{61.42\%} & 41.83\% & \textbf{62.07\%} & 69.58\% & 35.72\% \\
			Zhang et al. (2021)~\cite{Zhang.Y:et.al:2021:NAACL} & \textbf{53.11\%} & - & \textbf{46.86\%} & 55.63\% & 73.71\% & \textbf{45.90\%} \\
			\hline
			Average   & 48.63\% & 59.56\% & 41.36\% & 55.70\% & 73.39\% & 36.98\% \\
			\hline
			%%%%%%%%%%%%%%%%%%%%%%%%% Attention	
			\multicolumn{7}{c}{Attention Mechanism} \\
			\hline
			%			Zhang et al. (2016) ~\cite{Zhang.B:et.al:2016:arXiv} & - & - &  34.82\% & 53.70\% & 70.95\% & 36.16\% \\
			Liu \& Li (2016) ~\cite{Liu.Y:Li.S:2016:EMNLP} & 46.29\% & 57.17\%  & 36.70\% & 54.48\% & 70.43\% & 38.84\% \\
			Lan et al. (2017) ~\cite{Lan.M:et.al:2017:EMNLP} & {47.80\%} & 57.39\%  & 40.73\% & {58.96\%} & 72.47\% & 38.50\% \\
			Zhang et al. (2018) ~\cite{Zhang.B:et.al:2018:Elsevier} & - & - &  34.84\% & 54.11\% & 71.11\% & 34.11\% \\
			Guo et al. (2018) ~\cite{Guo.F:et.al:2018:COLING} & 47.59\% & 59.06\% & 40.35\% & 56.81\% & 72.11\% & 38.65\% \\
			Bai \& Zhao (2018) ~\cite{Bai.H:Zhao.H:2018:COLING} & 51.06\% & - & 47.85\% & 54.47\% & 70.60\% & 36.87\% \\
			Guo et al. (2020) ~\cite{Guo.F:et.al:2020:AAAI} & 47.90\% & 57.25\% & 43.92\% & 57.67\% & 73.45\% & 36.33\% \\
			Zhou et al. (2020)~\cite{Zhou.M:et.al:2020:IEEE}  & 48.75\%  & 59.18\% & 33.48\% & 52.19\% & 70.11\% & 39.21\% \\
			Ruan et al. (2020)~\cite{Ruan.H:et.al:2020:COLING} & - & - & {46.75\%} & {59.56\%} & {75.83\%} & {39.35\%} \\
			Li et al. (2020)~\cite{Li.X:et.al:2020:COLING} & - & - & {50.91\%} & 58.88\% & \underline{76.35\%} & {43.51\%} \\
			Liu et al. (2020) ~\cite{Liu.X:et.al:2020:IJCAI} & \textbf{63.39\%} & \textbf{69.06\%} & \underline{59.44\%} & \underline{60.98\%} & \textbf{77.66\%} & \underline{50.26\%} \\		
			Kishimoto et al. (2020) ~\cite{Kishimoto.Y:et.al:2020:LREC} & \underline{58.48\%} & \underline{65.26\%} & \textbf{75.46\%} & \textbf{73.01\%} & 72.86\% & \textbf{80.88\%} \\	
			Munir et al. (2021)~\cite{Munir.K:et.al:2021:IEEE} & {54.20\%} & {63.10\%} & {49.76\%} & 55.20\% & 73.10\% & 42.10\% \\
			\hline
			Average   & 51.72\% & 60.93\% & 46.68\% & 58.03\% & 73.01\% & 38.75\% \\
			\hline		
			%%%%%%%%%%%%%%%%%%%%%%%%% Arg-Pair	
			\multicolumn{7}{c}{Argument Pair Interaction} \\
			\hline		
			Chen et al. (2016a) ~\cite{Chen.J:et.al:2016:AAAI} & - & - & 30.21\% & 53.57\% & \textbf{80.90\%} & 20.24\% \\
			Chen et al. (2016b) ~\cite{Chen.J:et.al:2016:ACL} & - & - & 40.17\% & 54.76\% & \underline{80.62\%} & 31.32\% \\
			Li et al. (2016) ~\cite{Li.H:et.al:2016:NLPCC} & - & - & {41.91\%} & 54.72\% & 71.54\% & 34.78\% \\
			Lei et al. (2017) ~\cite{Lei.W:et.al:2017:IJCAI} & {46.46\%} & - & 40.47\% & {55.36\%} & {69.50\%} & {35.34\%} \\
			Varia et al. (2019) ~\cite{Varia.S:et.al:2019:SIGDIAL}  & \textbf{51.84\%} & \textbf{60.52\%} &  \underline{45.03\%} & \textbf{56.53\%} & 73.50\% & \textbf{46.15\%} \\
			He et al. (2020) ~\cite{He.R:et.al:2020:ACL} & \underline{51.24\%} & \underline{59.94\%} & \textbf{47.98\%} & \underline{55.62\%} & 69.37\% & \underline{38.94\%} \\
			\hline
			Average   & 49.85\% & 60.23\% & 40.96\% & 55.09\% & 74.24\% & 34.46\% \\
			\hline
	\end{tabular}}
\end{table*}

%%%%%%%%%%%%%%%%%%%%%%%%%%%%%%%%%%%%%%%%%%%%%%%%%%%%%%%%%%%%%%%%%%%%%%%%%%%%%%%%%%%%%
%%
%% Sec: Conclusion and Discussion
%%
%%%%%%%%%%%%%%%%%%%%%%%%%%%%%%%%%%%%%%%%%%%%%%%%%%%%%%%%%%%%%%%%%%%%%%%%%%%%%%%%%%%%%
\section{Conclusion and Discussion}\label{Sec:Conclusion}
Implicit discourse relation recognition is to detect relations and classify their senses in between arguments without explicit connectives. As a crucial task in the NLP field, the IDRR task has been intensively researched in the last decade.

\par
This article has presented a comprehensive survey for the IDRR task, including the task definitions, common datasets, solution approaches and performance comparisons. We have adopted the mostly researched PDTB corpus and its task definition to review three main groups of almost all solutions proposed in the literature. Those machine learning solutions manually design many features by domain experts to train classifiers; Those deep learning solutions design neural networks with different architectures to enable a kind of end-to-end relation recognition without manual feature construction. Both ML and DL solutions require large amounts of labeled training data; While semi-supervised approaches apply data expansion techniques to augment model training to deal with the data sparsity problem. Comparisons on the PDTB English dataset indicate that those sophiscated neural models can achieve the state-of-the-art performance.

\par
The challenge of the IDRR task mainly lies in the absence of explicit connectives in raw texts. If explicit connectives exists in raw texts, Pilter et al. (2008)~\cite{Pitler.E:et.al:2008:COLING} have reported as high as 93\% relation recognition accuracy; While the most advanced neural models can only achieve about 80\% recognition accuracy in the IDRR task. Lin et al.~\cite{Lin.Z:et.al:2009:EMNLP} summarized four reasons for the poor performance, namely, ambiguity between relations, inference, contextual modeling and world knowledge. These suggest that deeper semantic representations, external knowledge exploitations and more robust models are needed in future solutions. In what follows, we discuss possible future research directions for the IDRR task.

\subsection{Interaction-boosted representation learning}
Many neural models have been designed to learn the representation of an argument from its word sequence and use simple operations like concatenation, fully-connected layer and pooling function to capture argument-level interactions from argument representation for relation recognition. However, these neural models have not fully exploit word-level interactions of two arguments. Indeed, word-pairs have been shown as important features for machine learning-based relation classifiers. Besides word-pair interactions, external interactions may also need to be considered, like interactions with external lexicons, corpora, and knowledge bases, as such external resources can provide some background or side information for better understanding arguments. New neural models are needed to encode such word-level and external interactions into representation learning for implicit relation recognition.

\subsection{Synthetic Implicit Corpus Refinement}
The PDTB corpus has provided a gold labeled dataset for the IDRR task. However, its volume is still far behind high demands for training sophisticated neural models. Although simple approaches like removing explicit connectives in the explicit relation corpus can be applied to enrich implicit relation dataset, many explicit relation instances have been shown with much differences with implicit relation instances in real-world raw texts, which might not be appropriate to be directly used as gold labeled data. Recently, adversarial learning has been proposed to train a kind of discriminative classifier with erroneous or modified input data. It is worth of trying such adversarial learning framework to produce high-quality synthetic implicit relation corpus. Furthermore, other semi-supervised neural models also need to be investigated for joint data expansion and model training.

\subsection{Joint relation recognition and discourse parsing}
A discourse often contains multiple arguments’ relations, some explicit and some implicit. Yet the logical structure of multiple relations needs to be further analyzed to help understanding the whole discourse, which is also one of the discourse parsing tasks. Current relation recognition is generally modeled as an argument-pair classification problem; While individual arguments’ relations are next utilized for discourse parsing. However, arguments and their relations might be better understood against the whole discourse, especially together with other arguments’ relations. Although a few of machine learning solutions have utilized shared arguments as contextual features, they still followed the procedure of a single argument-pair classification. How to capture discourse-level information needs to be next investigated, like using preceding and following arguments and their relations. Furthermore, the design objective and operation structure of new neural models shall become an important research focus for joint relation recognition and discourse parsing.

\begin{acks}
	This work is supported in part by National Natural Science Foundation of China (Grant No: 62172167). The corresponding author is Bang Wang.
\end{acks}

\bibliographystyle{ACM-Reference-Format}
\bibliography{main}

\end{document}